\documentclass[acmsmall]{acmart}

\setcopyright{acmcopyright}
\copyrightyear{2023}
\acmYear{2023}
\acmDOI{XXXXXXX.XXXXXXX}

\acmJournal{TKDD}
\acmVolume{XX}
\acmNumber{XX}
\acmArticle{1}
\acmMonth{5}

\usepackage{amsfonts}
\usepackage{amsmath}
\usepackage{amstext}
\usepackage{amsthm}
\usepackage{bm}
\usepackage{booktabs}
\renewcommand{\arraystretch}{1}
\usepackage{enumitem}
\usepackage{graphicx}
\usepackage{multirow}
\usepackage{subfigure}
\usepackage{xcolor}

\newcommand{\specialcell}[2][c]{\begin{tabular}[#1]{@{}c@{}}#2\end{tabular}}

\newcommand{\model}{NETS-ImpGAN}
\newcommand{\modelp}{NETS-ImpGAN$^*$}
\newcommand{\framework}{ImpGAN}
\newcommand{\neuralnetwork}{GTAN}
\newcommand{\imputationgenerator}{GTA U-Net}

\newcommand{\gpred}{g_\text{pred}}
\newcommand{\gimp}{g_\text{imp}}

\newcommand{\bX}{\mathbf{X}}
\newcommand{\bM}{\mathbf{M}}
\newcommand{\bO}{\mathbf{O}}
\newcommand{\bZ}{\mathbf{Z}}
\newcommand{\bY}{\mathbf{Y}}
\newcommand{\bH}{\mathbf{H}}
\newcommand{\bQ}{\mathbf{Q}}
\newcommand{\bK}{\mathbf{K}}
\newcommand{\bV}{\mathbf{V}}
\newcommand{\bW}{\mathbf{W}}

\newcommand{\bomega}{\bm{\omega}}

\newcommand{\cG}{\mathcal{G}}
\newcommand{\cV}{\mathcal{V}}
\newcommand{\cE}{\mathcal{E}}
\newcommand{\cT}{\mathcal{T}}
\newcommand{\cD}{\mathcal{D}}
\newcommand{\cL}{\mathcal{L}}
\newcommand{\cF}{\mathcal{F}}

\newcommand{\R}{\mathbb{R}}
\newcommand{\E}{\mathbb{E}}

\begin{document}

\title{Networked Time Series Prediction with Incomplete Data via Generative Adversarial Network}

\author{Yichen Zhu}
\affiliation{%
  \institution{Shanghai Jiao Tong University}
  \city{Shanghai}
  \country{China}
}
\email{zyc_ieee@sjtu.edu.cn}

\author{Bo Jiang}
\authornotemark[1]
\thanks{*Bo Jiang is the corresponding author.}
\affiliation{%
  \institution{Shanghai Jiao Tong University}
  \city{Shanghai}
  \country{China}
}
\email{bjiang@sjtu.edu.cn}

\author{Haiming Jin}
\affiliation{%
  \institution{Shanghai Jiao Tong University}
  \city{Shanghai}
  \country{China}
}
\email{jinhaiming@sjtu.edu.cn}

\author{Mengtian Zhang}
\affiliation{%
  \institution{Shanghai Jiao Tong University}
  \city{Shanghai}
  \country{China}
}
\email{zhangmengtian@sjtu.edu.cn}

\author{Feng Gao}
\affiliation{%
  \institution{Zhejiang Lab}
  \city{Hangzhou}
  \country{China}
}
\email{gaof@zhejianglab.com}

\author{Jianqiang Huang}
\affiliation{%
  \institution{Alibaba Damo Academy}
  \city{Hangzhou}
  \country{China}
}
\email{jianqiang.jqh@gmail.com}

\author{Tao Lin}
\affiliation{%
  \institution{Communication University of China}
  \city{Beijing}
  \country{China}
}
\email{lintao@cuc.edu.cn}

\author{Xinbing Wang}
\affiliation{%
  \institution{Shanghai Jiao Tong University}
  \city{Shanghai}
  \country{China}
}
\email{xwang8@sjtu.edu.cn}

\renewcommand{\shortauthors}{Zhu et al.}

\begin{abstract}
  A \textit{networked time series (NETS)} is a family of time series on a given graph, one for each node. It has  a wide range of applications from intelligent transportation, environment monitoring to smart grid management. An important task in such applications is to predict the future values of a NETS based on its historical values and the underlying graph. Most existing methods require complete data for training. However, in real-world scenarios, it is not uncommon to have missing data due to sensor malfunction, incomplete sensing coverage, etc. In this paper, we study the problem of \textit{NETS prediction with incomplete data}. We propose {\model}, a novel deep learning framework that can be trained on incomplete data with missing values in both history and future. Furthermore, we propose  \textit{Graph Temporal Attention Networks}, which incorporate the attention mechanism to capture both inter-time series  and temporal correlations. We conduct extensive experiments on four real-world datasets under different missing patterns and missing rates. The experimental results show that {\model} outperforms existing methods, reducing the MAE by up to 25\%.
\end{abstract}




\maketitle

%%!TEX root = ../main.tex

\section{Introduction}\label{sec:introduction}

A networked time series (NETS) is a family of time series on a given graph, where each node is associated with a time series~\cite{cai2015fast}. Depending on the application, the underlying graph may encode spatial proximity, statistical dependency, or other contextual or structural information about the time series. As a versatile modeling tool, NETS has a wide range of applications from intelligent transportation, environment monitoring to smart grid management.

An important task in such applications is to predict the future values of a NETS based on its historical values and the underlying graph. This has been studied extensively and many prediction methods have been proposed in various contexts; see \cite{zhang2016dnn-based,zhang2017deep,yao2018deep,yao2019revisiting,li2018diffusion,yu2018spatio-temporal,wu2019graph,huang2020lsgcn,kuppannagari2021spatio-temporal,alcaraz202diffusion-based,geng2019spatiotemporal,bai2019stg2seq,ou2020spt-trellisnets,sun2021mobile,cai2015fast,zhong2021heterogeneous,wang2023traffic,cini2022filling} and references therein. Most of these methods use deep learning and require complete data for training \cite{zhang2016dnn-based,zhang2017deep,yao2018deep,yao2019revisiting,li2018diffusion,yu2018spatio-temporal,wu2019graph,huang2020lsgcn,kuppannagari2021spatio-temporal,alcaraz202diffusion-based,geng2019spatiotemporal,bai2019stg2seq,ou2020spt-trellisnets,sun2021mobile,zhong2021heterogeneous,wang2023traffic}. They typically use GCN \cite{kipf2017semi-supervised} to capture inter-time series correlations, and variants of CNN \cite{waibel1989phoneme} or RNN \cite{rumelhart1986learning} to capture temporal correlations. However, in real-world scenarios, it is not uncommon to have missing data due to sensor malfunction, incomplete sensing coverage, etc. Simply removing all samples with missing data could lead to low data efficiency as  demonstrated later in Section \ref{subsec:efficiency_study}, since some observed data will also be removed \cite{allison2001missing}. First imputing the incomplete data and then predicting with the imputed data can result in error accumulation, as shown in Section \ref{subsec:comparison_with_two-phase_prediction}. This motivates us to study the problem of \textit{NETS prediction with incomplete data}.

Several deep learning methods \cite{kuppannagari2021spatio-temporal,wu2021inductive,alcaraz202diffusion-based,zhong2021heterogeneous,wang2023traffic} can predict from incomplete history data. They take incomplete history as input and compute reconstruction loss on the complete predicted future. However, they require complete future data for supervision during training and do not provide a full solution to our problem.

DCMF \cite{cai2015fast} and GRIN \cite{cini2022filling} are two NETS imputation methods that do not require complete data for training. They can be used for prediction by treating the future as missing data \cite{cai2015facets}. DCMF is always fitted to a single sample, so it cannot benefit from training on multiple samples to learn more complex temporal dependencies  than allowed by its assumed linear system model. GRIN can be trained on multiple samples; however, it performs supervision only through the reconstruction loss on the observed values, with no direct control on the quality of the more important missing part. In addition, its bidirectional architecture may not be a perfect match for prediction, since there is no information passing from future to past.

MisGAN \cite{li2019learning} is a general distribution learning framework that can supervise the missing part with incomplete data. It can be adapted for NETS prediction by plugging in properly designed inner modules.  The framework first learns the joint distribution of the complete history and future, and then uses the learned complete data distribution to supervise prediction.  As later shown in Section \ref{subsubsec:comparison_with_misgan}, the imperfectly learned complete data distribution can lead to significant error accumulation.

To overcome the aforementioned problems, we present \textit{{\model}}, a novel deep learning framework for NETS prediction with incomplete data. {\model} consists of an outer framework called \textit{Imputation GAN ({\framework})}, and a collection of inner modules called \textit{Graph Temporal Attention Networks ({\neuralnetwork}s)}.

The outer framework {\framework} is a Generative Adversarial Net (GAN) \cite{goodfellow2014generative} for imputation. Conceptually, we can regard prediction as a special type of imputation as in \cite{cai2015facets}, where the future is considered as missing.  {\framework} aims to learn the conditional distribution of missing values given observed values. By supervision on distributions rather than on individual values as in GRIN, we can more readily use observed values in different incomplete samples to guide the prediction. More specifically, {\framework} has a generator that takes incomplete history and noise as input, and outputs complete samples including predicted future and completed history. The complete samples are then properly masked to generate new fake incomplete samples, which the  discriminator tries to distinguish from real incomplete samples. As such, {\framework} is a generic framework for imputation and may be of independent interest (Section \ref{subsubsec:imputation_performance}).  Note that real incomplete samples are  used directly to guide the imputation process, which largely avoids the error accumulation in MisGAN.

The inner {\neuralnetwork} modules specialize {\framework} to NETS prediction. As the implementations of the generators and discriminators of {\framework}, GTANs are designed to properly capture inter-time series correlations and temporal correlations. More specifically, Graph Attention Network~\cite{velickovic2018graph} is used to capture inter-time series correlations, and Multi-Head Self-Attention \cite{vaswani2017attention} and CNN \cite{waibel1989phoneme} are used to capture temporal correlations. We follow the common practice of filling missing values with random noise or constants, so that all samples have the same shape. As noted in \cite{zhong2021heterogeneous}, this may lead to inferior performance due to error accumulation. The incorporated attention mechanisms can help mitigate error accumulation by differentiating the observed values and filled values. Note that {\neuralnetwork}s can also be used as a standalone model for NETS prediction with complete data (Section \ref{subsubsec:two-phase_prediction}), or plugged into other frameworks such as MisGAN (Section \ref{subsubsec:comparison_with_misgan}).

To summarize, we make the following contributions.
\begin{itemize}[leftmargin=*]
    \item We propose {\model}, a novel deep learning framework for NETS prediction with incomplete data. The proposed framework can capture the complex dependencies from history to future with data that has missing values in both history and future.
    \item We propose {\neuralnetwork}s to capture the inter-time series correlations and temporal correlations of incomplete NETS. {\neuralnetwork}s mitigate error accumulation by incorporating  attention mechanisms.
	\item We conduct experiments on four real-world datasets under different missing patterns and missing rates. The results show that {\model} outperforms existing methods with up to 25\% reduction in prediction error.
\end{itemize}

The rest of the paper is organized as follows. Section \ref{sec:related_works} reviews the related work. Section \ref{sec:problem_formulation} formulates the problem. Section \ref{sec:methodology} presents the {\model} framework, followed by evaluation in Section \ref{sec:evaluation}. Section \ref{sec:conclusion} concludes the paper.

%%!TEX root = ../main.tex

\section{Related Work}\label{sec:related_works}

\textbf{NETS prediction}. The problem of NETS prediction has been studied extensively in various contexts. Early works \cite{zhang2016dnn-based,zhang2017deep} capture inter-time series correlations with CNN \cite{waibel1989phoneme} and temporal correlations by aggregating different timestamps with linear weighted sum. Some works \cite{yao2018deep,yao2019revisiting} then use LSTM \cite{hochreiter1997long} to capture non-linear temporal correlations. All these works use CNN and are limited to grid-structured graph. Many later works use GCN \cite{kipf2017semi-supervised} or its variants for the general graph setting. We only introduce the state-of-the-art methods below. DCRNN \cite{li2018diffusion} proposes diffusion-based GCN for inter-time series correlations and uses GRU \cite{cho2014learning} for temporal correlations. STGCN \cite{yu2018spatio-temporal} combines GCN and GLU \cite{dauphin2017language} to capture both inter-time series and temporal correlations. Graph WaveNet \cite{wu2019graph} captures inter-time series correlations by proposing a variant of GCN with adaptive adjacency matrix and uses WaveNet \cite{oord2016wavenet} to capture temporal correlations. LSGCN \cite{huang2020lsgcn} uses a variant of GCN with gated mechanism and GLU to capture inter-time series and temporal correlations, respectively. Some other works, for example \cite{geng2019spatiotemporal,bai2019stg2seq,ou2020spt-trellisnets,sun2021mobile}, additionally incorporate periodic or scenario-specific auxiliary information to improve the performance. The prediction of tensor-valued NETS has also been considered in \citet{jing2021network}, which proposes Tensor GCN and Tensor RNN. However, all these methods require complete history data as input.

Some works on NETS prediction can take incomplete history data as input. STGNN-DAE \cite{kuppannagari2021spatio-temporal} combines GCN and GLU in a denoising autoencoder for imputation. IGNNK \cite{wu2021inductive} generates random subgraphs and uses Diffusion GCN \cite{li2018diffusion} to learn the spatial message passing mechanism for imputation. STGNN-DAE and IGNNK can be used for prediction by treating the future as missing data. SSSDS4 \cite{alcaraz202diffusion-based} uses a generative framework based on conditional diffusion \cite{kong2021diffwave} and incorporates structured state space model \cite{gu2022efficiently} for long-term temporal correlations. RIHGCN \cite{zhong2021heterogeneous} is a deep learning framework for NETS prediction. It first uses GCN and LSTM to impute missing data in the history, and then uses the completed history to predict the future. The imputation and prediction steps are trained jointly to mitigate error accumulation. RIHGCN also uses multiple graphs, static and dynamic, to capture dynamic inter-time series correlation. GSTAE \cite{wang2023traffic} regard imputation and prediction as parallel tasks and train them sequentially to mitigate error accumulation. It combines GCN and GRU in an autoencoder for NETS prediction. However, all these methods require complete future data for training. 
 
There is limited literature on NETS prediction that allows incomplete input and does not require complete data for training. DCMF \cite{cai2015fast} is a matrix factorization method for missing data imputation of NETS. Its key assumptions are that a NETS and its associated graph have certain low-rank matrix representations, and that the temporal dynamics is described by a first-order linear system. Facets \cite{cai2015facets} extends DCMF to  tensor-valued NETS by using tensor decomposition instead of matrix factorization. NetDyna \cite{hairi2019netdyna} further considers the case where the underlying graph is also incomplete. All three models are fitted to a single sample with no future data. Without explicit learning to predict the future, their predictive power relies critically on the strong and potentially restrictive assumption of linear system model. S-MKKM \cite{gong2020a} incorporates a spatial multi-kernel clustering method into adaptive-weight non-negative matrix factorization for imputation of NETS. WDGTC \cite{li2020tensor} considers so-called weakly dependent modes in tensor completion. SD-ADMM \cite{meyers2023signal} decomposes a vector of time series into components with different characteristics, such as smooth, periodic, nonnegative, or sparse. Without a model for temporal dynamics, S-MKKM, WDGTC and SD-ADMM lacks the ability to predict the future. GRIN \cite{cini2022filling} combines bidirectional RNN \cite{schuster1997bidirectional} with GCN for NETS imputation. It computes reconstruction loss only on the observed part of the samples, and has no direct supervision on the missing part. SPIN \cite{marisca2022learning} uses attention mechanism along both the graph and temporal dimensions for NETS imputation, but it requires additional auxiliary information of temporal features and geographic location, which is not available in our problem.

In what follows, we also review the imputation methods for other data types that does not require complete data for training.

\textbf{Multiple time series imputation}. There are many methods for multiple time series imputation in the literature. They can also be used for prediction by treating the future as missing data. Among them, TRMF \cite{yu2016temporal}, BRITS \cite{cao2018brits}, E\textsuperscript{2}GAN \cite{luo2018multivariate,luo2019e2gan} and CSDI \cite{tashiro2021csdi} can use incomplete data for training. TRMF is also natively proposed for the prediction task, and it combines a novel regularization scheme along the temporal dimension with matrix factorization. BRITS imputes missing values using a bidirectional RNN with a temporal decay factor. It minimizes the reconstruction errors on observed values and forces consistency between the imputed values in both directions. E\textsuperscript{2}GAN is a GAN-based \cite{goodfellow2014generative} framework where the generator imputes missing values and the discriminator tries to distinguish imputed samples from samples with constant filled missing values. To capture temporal correlations of incomplete data, the generator and discriminator use a novel GRUI cell, which incorporates a temporal decay factor into GRU. CSDI uses a generative framework based on conditional diffusion \cite{alcaraz202diffusion-based} and incorporates Transformer layers \cite{vaswani2017attention} along both the temporal and feature dimensions. mSSA \cite{agarwal2022on} uses low order polynomials for trends, finite sum of harmonics for seasonality and linear time-invariant systems for imputation and prediction of incomplete multiple time series. All these methods take into account the correlations between all the time series in the imputation process, effectively treating multiple time series as a special NETS with a complete underlying graph. As such, they cannot exploit the additional information provided by the graph of a general NETS.

\textbf{General data imputation}. There are several imputation methods for general data. GAIN \cite{yoon2018gain} is an adaptation of GAN, where the generator imputes the missing entries, and the discriminator distinguishes between the observed and missing entries. MisGAN \cite{li2019learning} learns the complete data distribution from incomplete data and uses it to supervise the missing entries in imputation. Partial VAE \cite{ma2019eddi}, MIWAE \cite{mattei2019miwae} and P-BiGAN \cite{li2020learning} extend VAE \cite{kingma2014auto-encoding}, IWAE \cite{burda2016importance} and BiGAN \cite{donahue2017adversarial} respectively to learn the prior and posterior distributions of incomplete data given its latent representation and mask. They maximizes the likelihood of the observed entries and have no direct control on the missing entries. MIRACLE \cite{kyono2021miracle} is proposed as a regularization scheme that encourages the imputation to be consistent with the causal structure of data. It needs to be used jointly with other imputation methods as a refinement. GENIE \cite{dockhorn2022genie} is proposed as a denoising solver for the diffusion-based data perturbation process. Though the data perturbation process may not be consistent with the actual missing pattern of incomplete data, it still has the potential to be used for imputation. All these methods except GENIE can use incomplete data for training. These methods are evaluated on images with 2-D CNN-based models to capture spatial correlations of grid-structured data, and cannot be directly applied to general NETS. We adapt MisGAN to NETS and compare with it in Section \ref{subsubsec:comparison_with_misgan}.

Compared to existing methods, the proposed {\model} can directly supervise the missing part with only incomplete future data and adapt to different incomplete samples when capturing both the inter-time series correlations and temporal correlations of NETS.

%%!TEX root = ../main.tex

\section{Problem Formulation}\label{sec:problem_formulation}

\subsection{Background and Notation}\label{subsec:background_and_notation}

A networked time series (NETS) is a family of time series defined on a given graph. Let $\cG=(\cV,\cE)$ be an undirected graph with node set $\cV=\{1,2,\dots,V\}$ and edge set $\cE\subset\cV\times\cV$. A NETS on $\cG$ over timestamps $\cT=\{1,2,\cdots,T\}$ is $(\bX,\cG)$, where $\bX=(X_{v,t})\in\R^{V\times T}$ is a matrix whose entry $X_{v,t}$ is the value at timestamp $t\in\cT$ of the time series on node $v\in\cV$. Since we consider the case where the graph $\cG$ is fixed and known, we will refer to a NETS by $\bX$ for simplicity. It is understood that the underlying graph $\cG$ is given.

In the presence of missing data, only part of $\bX$ is observed. A binary mask $\bM=(M_{v,t})\in\{0,1\}^{V\times T}$ indicates which entries of $\bX$ are observed: $M_{v,t}=1$ if $X_{v,t}$ is observed, and $M_{v,t}=0$ if $X_{v,t}$ is missing. The complementary mask $\overline\bM\in\{0,1\}^{V\times T}$ is defined by $\overline{M}_{v,t}=1-M_{v,t}$, $\forall v,t$. With a slight abuse of notation, we also regard $\bM$ and $\overline\bM$ as the index sets of the observed and missing entries, so that the observed values are $\bX_\bM=\{X_{v,t}\mid(v,t)\in\bM\}$ and the missing values are $\bX_{\overline\bM}=\{X_{v,t}\mid(v,t)\in\overline\bM\}$. We consider the case where it is known which entries are observed, i.e., $\bM$ is known. Thus an incomplete data sample is given by $(\bX_{\bM},\bM)$. An incomplete dataset consists of $N$ such samples, denoted by $\cD=\{(\bX_{\bM^{(i)}}^{(i)},\bM^{(i)})\}_{i=1,2,\cdots,N}$.

Following \cite{little1986statistical}, we model the generative process of incomplete data as follows. A complete data sample $\bX$ is first drawn from the complete data distribution $p(\bX)$. Given $\bX$, a mask sample $\bM$ is then drawn from the conditional mask distribution $p(\bM\mid\bX)$. The resulted incomplete data sample $(\bX_{\bM},\bM)$ follows the distribution
\begin{equation*}
	p(\bX_\bM,\bM)=\int p(\bX)p(\bM\mid\bX)d\bX_{\overline\bM}\text{.}
\end{equation*}
We focus on the Missing Completely At Random (MCAR) case \cite{little1986statistical} where the mask $\bM$ is independent of the underlying complete data $\bX$, i.e., $p(\bM\mid\bX)=p(\bM)$.  Our proposed framework can be easily generalized to the Missing At Random (MAR) case \cite{little1986statistical} where $\bM$ only depends on the observed data $\bX_\bM$, i.e., $p(\bM\mid\bX)=p(\bM\mid\bX_\bM)$.

\subsection{Problem Statement}\label{subsec:problem_statement}

Given an incomplete history $(\bX_{\bM^h}^h,\bM^h)$ over $T^h$ timestamps, our task is to predict the future $\bX^f$ over the next $T^f$ timestamps. Specifically, we seek a prediction function $\gpred$ that takes $(\bX_{\bM^h}^h,\bM^h)\sim p(\bX_{\bM^h}^h,\bM^h)$ as input and outputs the predicted future $\hat\bX^f$.

We allow $\gpred$ to be random to accommodate multiple prediction~\cite{little1986statistical}, where multiple predicted samples are provided to reflect the uncertainty. We would like the predicted samples to follow the conditional distribution of the future given the incomplete history,
\begin{equation*}
 \hat\bX^f = \gpred(\bX_{\bM^h}^h,\bM^h)\sim p(\bX^f\mid\bX_{\bM^h}^h,\bM^h)\text{.}
\end{equation*}
When a single prediction is desired, we can use a summary statistic such as the mean of multiple predicted samples. Note $\gpred$ may depend on the underlying graph $\cG$, which is part of the input.

The predictor $\gpred$ will be trained on an incomplete dataset with missing values in both history and future. A sample in the dataset takes the form of $(\bX_{\bM^h}^h,\bM^h;\bX_{\bM^f}^f,\bM^f)$, where $(\bX_{\bM^f}^f,\bM^f)$ is the incomplete future.

%%!TEX root = ../main.tex

\section{Methodology}\label{sec:methodology}

We first introduce the proposed {\framework} framework in Section~\ref{subsec:framework}. Then we present in Section~\ref{subsec:graph_temporal_attention_networks} the detailed design of modules that specializes {\framework} to {\model}.

\subsection{The {\framework} Framework}\label{subsec:framework}

\begin{figure}[htbp]
	\centering
	\includegraphics[width=0.75\columnwidth]{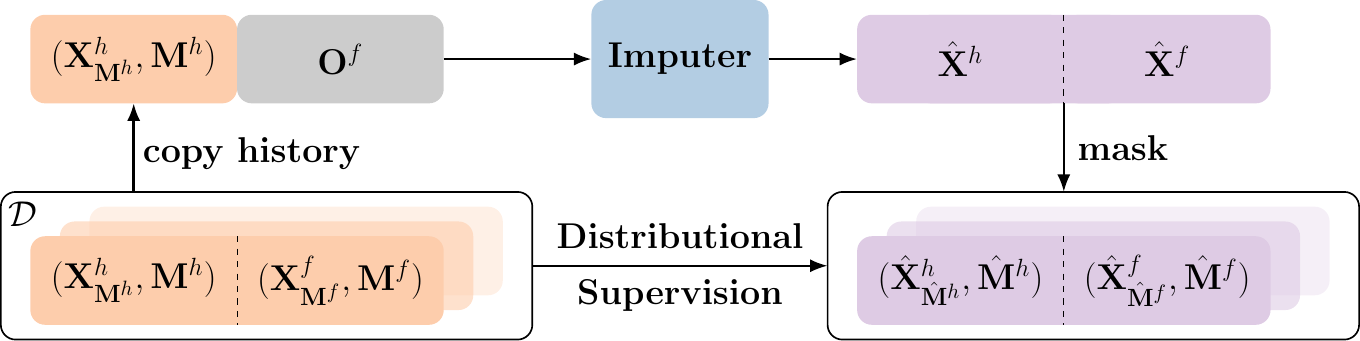}
	\caption{Supervision.}
	\Description{}
	\label{fig:supervision}
\end{figure}

A key challenge for training $\gpred$ is how to use the incomplete future $(\bX_{\bM^f}^f,\bM^f)$ to supervise the complete prediction $\hat\bX^f$. Supervising only on the observed future values can lead to inferior performance. We overcome this issue by supervising on the joint distribution of incomplete history and future. Specifically, we regard the prediction problem as a special type of imputation problem by treating the future as missing data. In Figure \ref{fig:supervision}, we feed incomplete history $(\bX_{\bM^h}^h,\bM^h)$ into the imputer and obtain the completed history and predicted future $(\hat\bX^h,\hat\bX^f)$. Then we mask $(\hat\bX^h,\hat\bX^f)$ by a generated mask to obtain an incomplete sample $(\hat\bX_{\hat\bM^h}^h,\hat\bM^h;\hat\bX_{\hat\bM^f}^f,\hat\bM^f)$, the distribution of which is supervised by that of the real data $(\bX_{\bM^h}^h,\bM^h;\bX_{\bM^f}^f,\bM^f)$.

To simplify the notation, we will use $(\bX_\bM,\bM)$ to denote a full sample of length $T=T^h+T^f$, consisting of both history and future, where $\bX=\bX^h\|\bX^f$, $\bM=\bM^h\|\bM^f$, with $\|$ denoting concatenation in the temporal dimension. By introducing a new mask $\bM^*=\bM^h\|\bO^f$, where $\bO^f$ is an all-zero mask of size $V\times T^f$, we can rewrite the input to the imputer as $(\bX_{\bM^*},\bM^*)$.
The output of the imputer is denoted by $\hat\bX=\hat\bX^h\|\hat\bX^f$. The imputed values $\hat\bX_{\overline{\bM^*}}$ consist of both the imputed history $\hat\bX_{\overline\bM^h}^h$ and the predicted future $\hat\bX^f$. Thus instead of learning $\gpred$, we can learn a random function $\gimp$ such that
\begin{equation*}
	\hat\bX_{\overline{\bM^*}}=\gimp(\bX_{\bM^*},\bM^*)\sim p(\bX_{\overline{\bM^*}}\mid\bX_{\bM^*},\bM^*)\text{,}
\end{equation*}
and then extract $\hat \bX^f$. Using the reparameterization trick, we can learn a deterministic function, still denoted by $\gimp$, that takes some random noise $\bZ$ as an additional input such that
\begin{equation*}
	\hat\bX_{\overline{\bM^*}}=\gimp(\bX_{\bM^*},\bM^*,\bZ)\sim p(\bX_{\overline{\bM^*}}\mid\bX_{\bM^*},\bM^*)\text{.}
\end{equation*}

\begin{figure}[t]
	\centering
	\includegraphics[width=0.65\columnwidth]{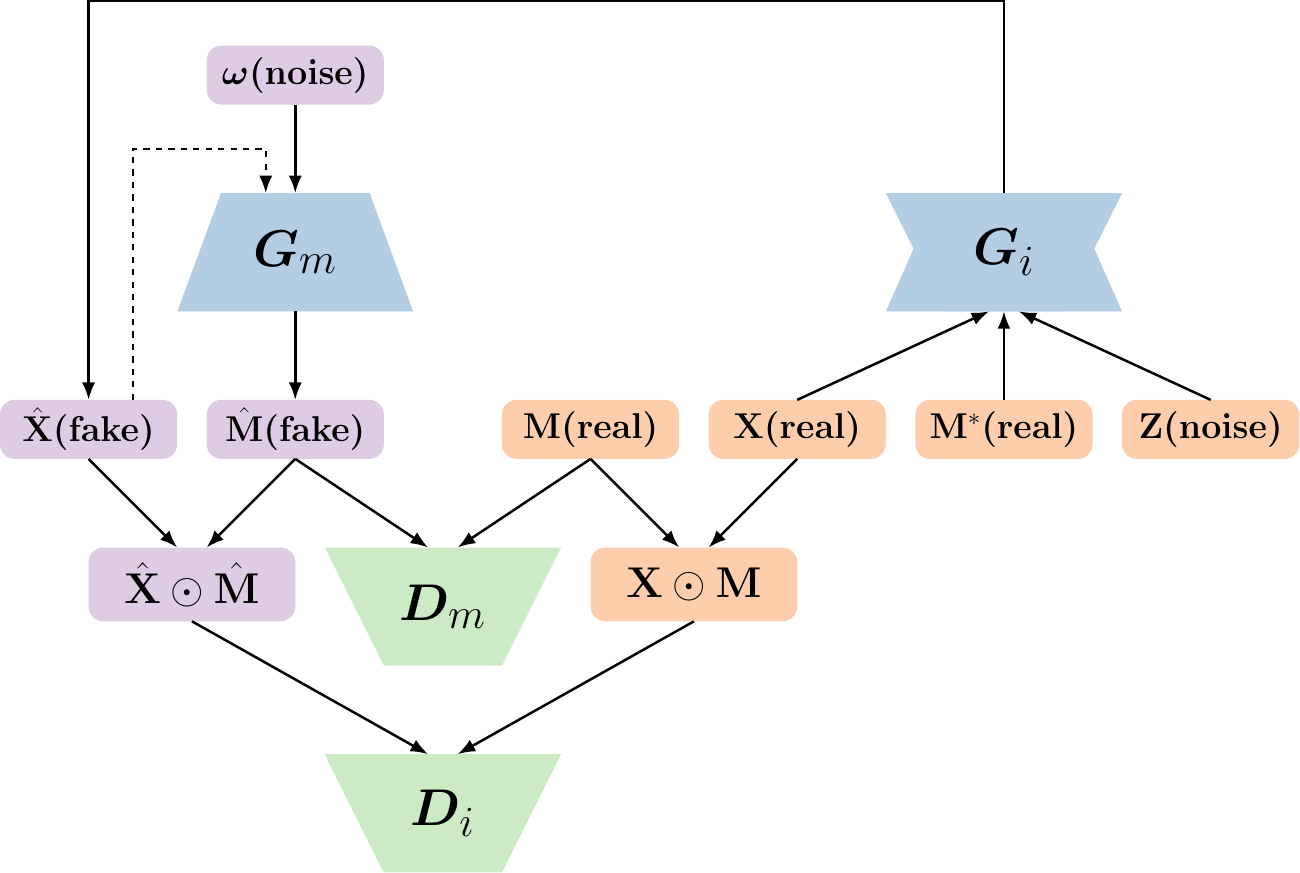}
	\caption{Architecture of {\framework}.}
	\Description{}
	\label{fig:framework}
\end{figure}

We adopt Generative Adversarial Net (GAN) \cite{goodfellow2014generative} to learn $\gimp$ for its high sampling efficiency and propose a deep learning framework named \textit{Imputation GAN ({\framework})}, as shown in Figure \ref{fig:framework}.
The function $\gimp$ is realized by the imputation generator $G_i(\bX,\bM^*,\bZ)$, where $\bM^*=\bM^h\|\bO^f$ is the mask indicating that the entire future is missing. The construction of $G_i$ is
\begin{equation*}
	G_i(\bX,\bM^*,\bZ)=\bX\odot\bM^*+\hat{G}_i(\bX\odot\bM^*+\bZ\odot\overline{\bM^*})\odot\overline{\bM^*}\text{,}
\end{equation*}
where $\bZ\sim p_\bZ$ is a random noise of size $V\times T$, $\odot$ is element-wise multiplication, and $\hat{G}_i$ is a function whose input and output are both of size $V\times T$. In this paper, $\hat{G}_i$ is implemented by the neural network to be introduced in Section \ref{subsec:graph_temporal_attention_networks}. Note that $G_i$ depends on $\bX$ only through the masked form $\bX\odot \bM^*$, so the true input to $G_i$ is actually $\bX_{\bM^*}$, and the missing data $\bX_{\overline{\bM^*}}$ is never needed. The masking by $\bM^*$ outside of $\hat{G}_i$ is used to retain the observed data.
If $G_i$ successfully captures the conditional distribution $p(\bX_{\overline{\bM^*}}\mid\bX_{\bM^*},\bM^*)$, then the output $\hat\bX$ will have the distribution $p(\bX)$ of $\bX$.

In order to train $G_i$, we need to use the incomplete data $(\bX_\bM,\bM)$ in $\cD$ to supervise the output $\hat\bX$. This is done by re-masking $\hat\bX$ back to an incomplete sample, which we then have the imputation discriminator $D_i$ try to distinguish from the real incomplete data.
More specifically, we use a standard GAN $(G_m(\bomega),D_m(\bM))$ to learn the mask distribution $p(\bM)$, as the mask $\bM=\bM^h\|\bM^f$ is fully observed. We then generate a mask $\hat\bM$ using $G_m$ and obtain the corresponding incomplete sample $(\hat\bX_{\hat\bM},\hat\bM)$, which $D_i$ tries to distinguish from real samples $(\bX_\bM, \bM)$ in $\cD$.
Since neural networks generally take arrays of fixed shape as input,
we fill missing entries with zeros and have $D_i$ distinguish between $\bX\odot\bM$ and $\hat\bX\odot\hat\bM$ instead. Note that the input
$\bX\odot\bM=(\bX_{\bM^h}^h\odot\bM^h)\|(\bX_{\bM^f}^f\odot\bM^f)$
 into $D_i$
includes partially observed future, while the input
$\bX\odot\bM^*=(\bX_{\bM^h}^h\odot\bM^*)\|\bO^f$
 into $G_i$ includes no future values. In this way, we are able to use only incomplete future to supervise prediction.

An alternative way of training $G_i$ is provided in MisGAN~\cite{li2019learning}, which learns the complete data distribution $p(\bX)$ from $(\bX_\bM,\bM)$ and uses the learned distribution $\hat p(\bX)$ to supervise imputation. However, as we will be see in Section \ref{subsubsec:comparison_with_misgan}, regarding the imperfectly learned $\hat p(\bX)$ as ground truth will result in error accumulation, which {\framework} avoids by directly using incomplete data for supervision. The imperfectly learned mask distribution $\hat p(\bM)$ could also cause error accumulation, but we will see in Section \ref{subsubsec:comparison_with_misgan} that this effect is negligible, as $p(\bM)$ is easy to learn.

Following the Wasserstein GAN \cite{arjovsky2017wasserstein} formulation, we define the training objectives by
\[
    \min\limits_{G_m}\max\limits_{D_m\in\cF_m}\cL_m(D_m,G_m)\text{,}\quad\min\limits_{G_i}\max\limits_{D_i\in\cF_i}\cL_i(D_i,G_i,G_m)\text{,}
\]
where $\cF_m$ and $\cF_i$ are the classes of $1$-Lipschitz functions for $D_m$ and $D_i$, respectively. The loss functions are
\[
	\begin{aligned}
		\cL_m(D_m,G_m)=&  \;\E[D_m(\bM)]-\E[D_m(G_m(\bomega))]\text{,}\\
		\cL_i(D_i,G_i,G_m) = & \;\E[D_i(\bX\odot\bM)]-\E[D_i(G_i(\bX,\bM^*,\bZ)\odot G_m(\bomega))]  \\
		  & +\beta\E\|\hat{G}_i(\bX,\bM^*,\bZ)\odot\bM^*-\bX\odot\bM^*\|_1\text{,}
	\end{aligned}
\]
where the expectations are taken over $(\bX,\bM)\sim p_\cD$, $\bomega\sim p_{\bomega}$ and $\bZ\sim p_\bZ$, $p_\cD$ is the underlying distribution of $\cD$, $\|\cdot\|_1$ is the $L_1$-norm, and $\beta$ is the trade-off parameter for $L_1$-norm. The $L_1$-norm term is the reconstruction loss for the output of $\hat{G}_i$, which is added to force the output of $\hat{G}_i$ to have the same observed history as the input.

Since we focus on the MCAR case, the mask generator $G_m(\bomega)$ only takes random noise $\bomega$ as input. To generalize to the MAR case, $G_m$ can take $\hat\bX$ as additional input to model the dependence of mask $\bM$ on the underlying $\bX$, as indicated by the dotted line in Figure \ref{fig:framework}.

{\framework} can also be turned into a generic imputation framework of independent interest. Recall that the data $\bX\odot \bM^*$ fed into $G_i$ include no future values and differ from the data $\bX\odot \bM$ fed into $D_i$. This is a consequence of the prediction task. If we feed the same data $\bX\odot \bM$ into both $G_i$ and $D_i$, {\framework} will then become a generic imputation framework that can be combined with different designs for the generators and discriminators. For the prediction task, the discrepancy between $\bX\odot \bM^*$ and $\bX\odot \bM$ requires $G_i$ to be able to capture temporal correlations at the minimum.

\subsection{Graph Temporal Attention Networks}\label{subsec:graph_temporal_attention_networks}

\subsubsection{{\imputationgenerator}}

We specialize {\framework} to {\model} by designing \textit{Graph Temporal Attention Networks ({\neuralnetwork}s)} for the generators and discriminators that capture both inter-time series and temporal correlations of incomplete NETS. This section presents the \textit{Graph Temporal Attention U-Net ({\imputationgenerator})} that we propose for the imputation generator $G_i$. The mask generator and the discriminators have similar building blocks; see Section \ref{subsubsec:mask_generator_and_discriminators} for their detailed architectures.

\begin{figure}[t]
	\centering
	\includegraphics[width=0.6\columnwidth]{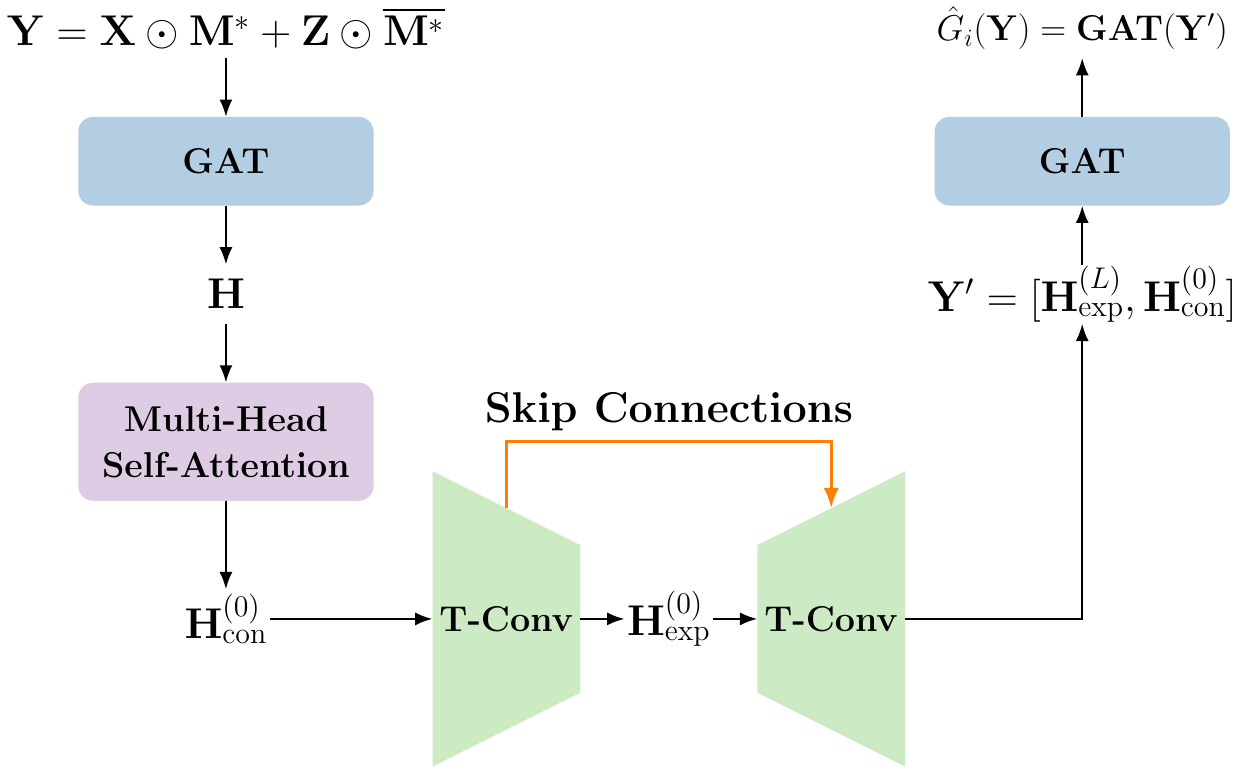}
	\caption{Architecture of {\imputationgenerator}.}
	\Description{}
	\label{fig:imputation_generator}
\end{figure}

Figure \ref{fig:imputation_generator} shows the architecture of {\imputationgenerator}.
We use Graph Attention Network (GAT) \cite{velickovic2018graph} to capture the inter-time series correlations, and we use Multi-Head Self-Attention \cite{vaswani2017attention} and Temporal Convolution (T-Conv) to capture global and local temporal correlations, respectively.
The input is $\bY = \bX\odot\bM^*+\bZ\odot\overline{\bM^*}$, consisting of both observed values and random noise. As noted in \cite{zhong2021heterogeneous}, this may lead to inferior performance due to error accumulation. The incorporated attention mechanisms can help mitigate error accumulation by differentiating the observed  and filled values.

Similar to U-Net \cite{ronneberger2015u-net}, {\imputationgenerator} has a U-shaped structure, where the contractive encoding path (left side) first extracts a lower-dimensional representation of the input incomplete data and the expansive decoding path (right side) then recovers the complete data from this representation. Along the encoding path, the input first goes through a GAT layer, then a Multi-Head Self-Attention layer and finally a stack of T-Conv layers. Along the decoding path, data goes through a stack of T-Conv layers and then a GAT layer.

\textbf{Graph Attention Layer.} The GAT layer on the contractive path takes $\bY$ 
as input and outputs $\bH=(H_{i,t})=\mathrm{GAT}(\bY)\in\R^{V\times T}$  given by
\begin{equation}\label{equ:gat}
	H_{i,t}=\sigma\left(\alpha_{i,i,t}\theta Y_{i,t}+\sum\limits_{j\in\mathcal{N}(i)}\alpha_{i,j,t}\theta Y_{j,t}\right),\quad i\in\cV,t\in\cT\text{,}
\end{equation}
where $\mathcal{N}(i)$ is the set of one-hop neighbors of node $i$, and $\theta\in\R$ is a training parameter, and $\sigma$ is an activation function. The attention coefficient $\alpha_{i,j,t}$ is given by
\begin{equation}\label{equ:attention_coefficient}
	\alpha_{i,j,t}=\frac{\exp{(\mathrm{LeakyReLU}(\mathbf{a}^\top[\theta Y_{i,t}\|\theta Y_{j,t}]))}}{\sum_{k\in\mathcal{N}(i)\cup\{i\}}\exp{(\mathrm{LeakyReLU}(\mathbf{a}^\top[\theta Y_{i,t}\|\theta Y_{k,t}]))}}\text{,}
\end{equation}
where $\|$ is concatenation, and $\mathbf{a}\in\R^2$ is a training parameter.

\textbf{Multi-Head Self-Attention Layer.} Multi-Head Self-Attention captures the temporal correlations from a global view of all the timestamps. The input to the Multi-Head Self-Attention layer is $\bH$. Multi-Head Self-Attention consists of multiple heads. For the $i$-th head, define query $\bQ_i=\bH^\top\bW^Q_i\in\R^{T\times d_k}$, key $\bK_i=\bH^\top\bW^K_i\in\R^{T\times d_k}$ and value $\bV_i=\bH^\top\bW^V_i\in\R^{T\times d_v}$, where $\bW^Q_i$, $\bW^K_i$ and $\bW^V_i$ are training parameters. The output of the $i$-th head is
\begin{equation*}
	\mathrm{head}_i=\mathrm{Softmax}\left(\frac{\bQ_i\bK_i^\top}{\sqrt{d_k}}\right)\bV_i\text{.}
\end{equation*}
The output of Multi-Head Self-Attention is then
\begin{equation*}
	\mathrm{MultiHead}=\bW^O\mathrm{Concat}(\mathrm{head}_1,\cdots,\mathrm{head}_h)^\top\text{,}
\end{equation*}
where $\mathrm{Concat}(\cdot)$ is concatenation along the second dimension, and $\bW^O\in\R^{V\times hd_v}$ is a training parameter.

\textbf{Temporal Convolution Layers.} Temporal Convolution (T-Conv) captures the temporal correlations from a local view of the neighboring timestamps, which complements Multi-Head Self-Attention. The convolutional structure also avoids error accumulation, which recurrent structures potentially suffer from. This leads to higher stability in multi-step prediction, as will be demonstrated in Section \ref{subsec:prediction_performance}.

We use 1-D CNN \cite{waibel1989phoneme} to halve and double the temporal dimensions along the contractive path and the expansive path, respectively. There are $L$ layers along each path. 
Let $\bH_{\mathrm{con}}^{(0)}\in \R^{V\times T}$ denote the output of Multi-Head Self-Attention, and $\bH_{\mathrm{con}}^{(l)}\in \R^{V\times T/2^l}$  that of the $l$-th T-Conv layer along the contractive path.
For $l=1,\dots,L$,
\begin{equation*}
	\bH_{\mathrm{con}}^{(l)}=\sigma\left(\mathrm{BN}(\mathrm{Conv1d}(\bH_{\mathrm{con}}^{(l-1)},\mathrm{stride}=2))\right)\text{,}
\end{equation*}
where  $\mathrm{Conv1d}(\cdot,\mathrm{stride}=2)$ is a 1-D CNN with stride 2, and $\mathrm{BN}(\cdot)$ is batch normalization.

The output $\bH_{\mathrm{con}}^{(L)}$ of the contractive path is sent to the expansive path. Let $\bH_{\mathrm{exp}}^{(0)}=\bH_{\mathrm{con}}^{(L)}$ and denote by $\bH_{\mathrm{exp}}^{(l)}\in\R^{V\times T/2^{L-l}}$ the output of the $l$-th T-Conv layer along the expansive path. For $l=1,\dots,L$,
\begin{equation*}
	\bH_{\mathrm{exp}}^{(l)}=\sigma\left(\mathrm{BN}(\mathrm{DeConv1d}([\bH_{\mathrm{exp}}^{(l-1)},\bH_{\mathrm{con}}^{(L-l+1)}],\mathrm{stride}=2))\right)\text{,}
\end{equation*}
where $\mathrm{DeConv1d}(\cdot,\mathrm{stride}=2)$ is a 1-D de-convolution with stride 2, and $[\cdot,\cdot]$ denotes concatenation along the feature dimension.

We have also followed U-Net \cite{ronneberger2015u-net} to add skip connections as indicated by the orange arrow in Figure \ref{fig:imputation_generator}. They are used to preserve information at the border, i.e. at the starting and ending timestamps, since such information tends to be lost in convolution.

The expansive path is followed by a final GAT layer, which takes as input $\bY^{\prime}=[\bH_{\mathrm{exp}}^{(L)},\bH_{\mathrm{con}}^{(0)}]\in\R^{V\times T\times2}$, and outputs
\begin{equation*}
	\hat{G}_i(\bY)=\mathrm{GAT}(\bY^{\prime})\text{.}
\end{equation*}
The GAT layer has a similar structure as in \eqref{equ:gat} and \eqref{equ:attention_coefficient}, except that the training parameter $\theta\in\R$ is replaced by $\bm{\Theta}\in\R^{1\times 2}$, since $\bY^\prime$ has 2-D feature, but the output $\mathrm{GAT}(\bY^\prime)\in\R^{V\times T}$ has only 1-D feature.

\subsubsection{Mask Generator and the Discriminators}\label{subsubsec:mask_generator_and_discriminators}

Mask generator $G_m$, mask discriminator $D_m$ and imputation discriminator $D_i$ have similar building blocks to those of {\imputationgenerator}.

\begin{figure}[t]
% 	\centering
	\subfigure[Mask Generator]{
% 		\begin{minipage}[t]{.45\columnwidth}
% 			\centering
			\includegraphics[width=.43\textwidth]{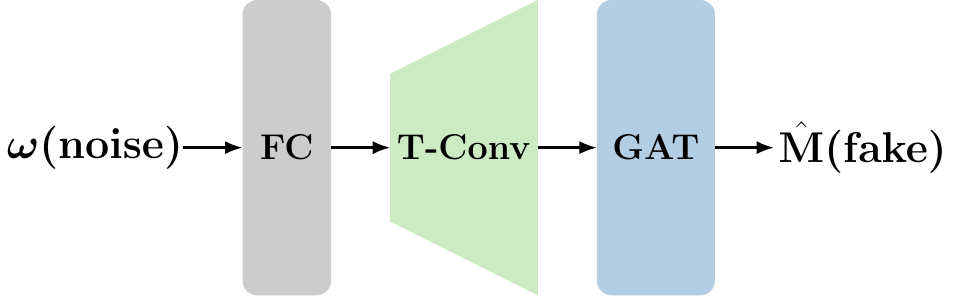}
			\label{fig:mask_generator}
% 		\end{minipage}
	}
	\hspace{1em}
	\subfigure[Discriminators]{
% 		\begin{minipage}[t]{.5\columnwidth}
			\centering
			\includegraphics[width=.5\textwidth]{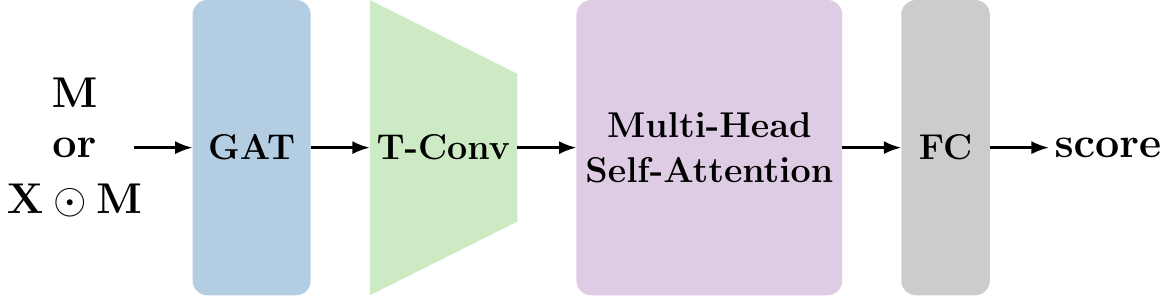}
			\label{fig:discriminators}
% 		\end{minipage}
	}
	\caption{Architecture of mask generator and the discriminators.}
	\Description{}
\end{figure}

The architecture of $G_m$ is shown in Figure \ref{fig:mask_generator}. The random noise $\bomega$ first goes through a Fully Connected (FC) layer, which reshapes $\bomega$ to the desired shape. Then the reshaped data goes through $L$ layers of T-Conv and a GAT layer, which is the same as the expansive path of {\imputationgenerator} without skip connections.

The discriminators $D_m$ and $D_i$ have the same architecture shown in Figure \ref{fig:discriminators}. The input goes through a GAT layer, a Multi-Head Self-Attention layer and $L$ layers of T-Conv, which is the same as the contractive path of {\imputationgenerator} but in the reverse order. This is followed by a Fully Connected (FC) layer, whose output is a scalar score that indicates how real the input is.

%%!TEX root = ../main.tex

\section{Evaluation}\label{sec:evaluation}

We introduce the experimental setup in Section \ref{subsec:experimental_setup}. Section \ref{subsec:prediction_performance} presents the prediction performance of {\model}. Section \ref{subsec:comparison_with_two-phase_prediction} gives the comparison with two-phase methods that first impute incomplete data and then use the imputed data to train prediction models. Section \ref{subsec:efficacy_study} gives the efficacy study. Section \ref{subsec:efficiency_study} gives the efficiency study.

\subsection{Experimental Setup}\label{subsec:experimental_setup}

\subsubsection{Datasets}\label{subsubsec:datasets}

We evaluate {\model} on the following four real-world datasets.

\textbf{Metro Passenger Flow (Metro)} \cite{tianchi2019tianchi} is a collection of passenger outbound flows from 81 metro stations in Hangzhou, China, within every 10 minutes in January 2019. This dataset is collected from automated fare collection records. In the underlying graph, nodes represent the metro stations, and edges are constructed based on pairwise transition probability between stations. In the presence of missing data, we first delete the automated fare collection records that correspond to the missing part of all the incomplete samples and then follow \citet{ou2020spt-trellisnets} to compute the transition probability matrix with the remaining records. A pair of nodes are connected by an undirected edge if at least one of the transition probabilities between them is no less than 0.02.

\textbf{Air Quality Index (Air)} \cite{microsoft2012urban} is a collection of air quality records from 437 monitoring stations in China within every hour from May 2014 to April 2015. Same as \citet{cini2022filling}, we focus on the PM2.5 pollutant. In the underlying graph, nodes represent the monitoring stations, and edges are constructed based on pairwise geographic distance between stations. We follow \citet{cini2022filling} that processes the distance with threshold Gaussian kernel. A pair of nodes are connected by an undirected edge if the processed distance between them is no less than 0.1.

\textbf{Electricity Consumption (Electricity)} \cite{irish2016issda} is a collection of electricity consumption records from 485 smart meters in Ireland within every 30 minutes from 2009 to 2010. In the underlying graph, nodes represent the smart meters, and edges are constructed based on the similarity between smart meters. We follow \citet{cini2022filling} that first computes a similarity matrix under correntropy and then builds a $k$-nearest neighbor graph where $k=10$.

All the above three datasets consist of complete samples. We generate  incomplete datasets by introducing missing values according to the missing patterns in Section \ref{subsubsec:missing_patterns}.

\textbf{Traffic Speed (Speed)} \cite{didi2018city} is a collection of traffic speed records from 1343 road segments in Chengdu, China, within every hour in 2018. In the underlying graph, nodes represent the road segments, and edges are constructed based on the adjacency of road segments. This dataset is naturally incomplete with an average missing rate of 30\%.

\subsubsection{Missing Patterns}\label{subsubsec:missing_patterns}

We follow previous works \cite{cao2018brits,luo2018multivariate,luo2019e2gan,tashiro2021csdi,yoon2018gain,li2019learning,ma2019eddi,mattei2019miwae,li2020learning} to evaluate the  Random pattern given below and also consider the more general MV pattern. The missing rate, which is denoted by $r$, is set to be low, medium and high at 25\%, 50\% and 75\%, respectively, and also very low at 2\%, 4\%, 6\%, 8\% and 10\%.
\begin{itemize}[leftmargin=*]
	\item \textbf{Random}. In each sample, values on all nodes at all timestamps are independently randomly missing with probability $r$.
	\item \textbf{Multiple Block of Variable Shape (MV)}. In each sample, multiple blocks are missing. In each block, values on $N_v$ nodes at randomly selected $N_t$ consecutive timestamps are missing, where $N_v$ and $N_t$ are uniformly randomly drawn from $[l_v,u_v]$ and $[l_t,u_t]$. The nodes are selected by breadth-first search from a random node until the desired number of nodes have been traversed. The number of blocks is $\lfloor\tfrac{VTr}{(l_v+u_v)(l_t+u_t)/4}\rfloor$. In this paper, we set $l_v=1$, $u_v=7$, $l_t=1$ and $u_t=3$. The blocks are selected independently, so the missing rate may not be exactly $r$ due to possible overlapping of the blocks.
\end{itemize}

\begin{table}[t]
	\centering
	\caption{Differences between {\model} and prediction baselines.}
	\label{tbl:prediction_difference}
	\renewcommand{\arraystretch}{0.9}
	\begin{tabular}{@{}lccc@{}}
		\toprule
		& \specialcell{{Supervision with Incomplete Future}} & {Graph} & {Temporal} \\ \midrule
		{Mean}						& $\times$	& $\times$	& $\times$ \\
		{TLE}						& $\times$	& $\times$	& $\surd$ \\
		{LO}						& $\times$	& $\times$	& $\surd$ \\
		{DCMF}						& $\times$	& $\surd$	& $\surd$ \\
		{TRMF}						& $\times$	& $\times$	& $\surd$ \\
		{BRITS}						& $\times$	& $\times$	& $\surd$ \\
		{E\textsuperscript{2}GAN}	& $\surd$	& $\times$	& $\surd$ \\
		{GRIN}						& $\surd$	& $\surd$	& $\surd$ \\
		{CSDI}						& $\times$	& $\times$	& $\surd$ \\
		{mSSA}						& $\times$	& $\times$	& $\surd$ \\
		{{\model}}					& $\surd$	& $\surd$	& $\surd$ \\
		\bottomrule
	\end{tabular}
\end{table}

\subsubsection{Baselines}\label{subsubsec:baselines}

We compare {\model} with the following prediction baselines: (1) simple methods including Mean, TLE and LO, and (2) state-of-the-art methods including DCMF \cite{cai2015fast}, TRMF \cite{yu2016temporal}, BRITS \cite{cao2018brits}, E\textsuperscript{2}GAN \cite{luo2018multivariate,luo2019e2gan}, GRIN \cite{cini2022filling}, CSDI \cite{tashiro2021csdi} and mSSA \cite{agarwal2022on}. For a fair comparison, we focus on methods that do not require complete data for training. Those that require complete data for training will be compared in Section \ref{subsec:comparison_with_two-phase_prediction}. Mean, TLE and LO are described as follows.
\begin{itemize}[leftmargin=*]
	\item \textbf{Mean}. Mean predicts the missing entries with the mean of observed entries in the same sample.
	\item \textbf{Temporal Linear Extrapolation (TLE)}. TLE predicts a missing entry with the linear function that passes through its last two observed historical entries on the same node in the same sample. Mean is used if the required entries are missing.
	\item \textbf{Last Observation (LO)}. LO predicts a missing entry with the last observed historical entry on the same node in the same sample. Mean is used if the required entries are missing.
\end{itemize}
For DCMF, BRITS, E\textsuperscript{2}GAN, GRIN and CSDI, we use them for prediction by treating the future as missing data. For all baselines, we will report the better of the results refined by MIRACLE \cite{kyono2021miracle} or not. We will report our performance both with and without the refinement. Note that Facets \cite{cai2015facets} and NetDyna \cite{hairi2019netdyna} are reduced to DCMF in our problem, so we only select DCMF as baseline.

Table \ref{tbl:prediction_difference} summarizes the differences between {\model} and the prediction baselines w.r.t. whether they can (1) supervise with incomplete future, (2) exploit the underlying graph and (3) capture temporal correlations.

\subsubsection{Evaluation Metrics}\label{subsubsec:evaluation_metrics}

We use Mean Absolute Error (MAE), Root Mean Square Error (RMSE) and Mean Absolute Percentage Error (MAPE) as evaluation metrics. Since we learn a distribution of the future, we have the flexibility to optimize the prediction results according to a given metric. Specifically, for MAE and RMSE, we use the sample medium and mean, respectively, of multiple samples generated by the imputation generator; for MAPE, we solve an empirical MAPE minimization problem by quantile regression \cite{koenker2001quantile}. Since DCMF, E\textsuperscript{2}GAN and CSDI also have distributions, we do the same optimization for them. The metrics are computed over all $T^f$ future timestamps unless stated otherwise. We will mostly report the results for MAE due to space limit, but the other results are similar.

\subsubsection{Implementation Details}\label{subsubsec:implementation_details}

Each generated sample has $T=16$ timestamps; the first 8 constitute  history, and the last 8  future. Since the ranges of values vary greatly across nodes, we apply Min-Max normalization to scale the data to $[-1,1]$ for each node before they are fed into the neural networks. The predicted or imputed values are then rescaled back to the original range.

The proposed {\model} is implemented with PyTorch. All Multi-Head Self-Attention modules have $h=3$ heads. The contractive and expansive paths have $L=3$ T-Conv layers, respectively. All T-Conv modules have kernel of size $3\times1$ and stride of 2, and the number of channels doubles along the contractive path and halves along the expansive path. The noise $\bomega$ fed into the mask generator $G_m$ is a 128-dimensional standard Gaussian random vector. For activation functions, we use  $\mathtt{LeakyReLU}$ throughout, except for the final GAT layer of $G_i$ and $G_m$, where $\mathtt{tanh}$ and $\mathtt{hardtanh}$ are used, respectively, in order to make the range of the generated data and mask consistent with that of the normalized real data and real mask.

We use 90\% of the samples for training, 5\% for validation and 5\% for testing. The model is trained for 1000 epochs with batch size 64. We follow the common practice that alternatively optimizes the discriminators for 5 epochs and the generators for 1 epoch. We use the Adam optimizer with learning rate 0.0001. The trade-off parameter $\beta$ in Section~\ref{subsec:framework} is set to 10.  In the testing phase, we randomly generate 10 samples by $G_i$ to compute the optimal prediction as specified in Section \ref{subsubsec:evaluation_metrics}.

\begin{table}[t]
	\centering
	\caption{Prediction performance on Metro, Air, Electricity under 25\% missing rate. The results of the baselines are the better of those refined by MIRACLE or not, and the results of {\model}$^\dag$ are refined by MIRACLE. \textbf{Bold}, \underline{underline} and superscript $^*$ indicate the best, second best and third best in each column. These marks will also be used in Table \ref{tbl:imputation_performance}.}
	\label{tbl:prediction_performance}
	\renewcommand{\arraystretch}{0.95}
	\subtable[Random]{
		\begin{tabular}{@{}lrrrrrrrrr@{}}
			\toprule
			\multirow{2.5}*{\textbf{Method}} & \multicolumn{3}{c}{\textbf{Metro}} & \multicolumn{3}{c}{\textbf{Air}} & \multicolumn{3}{c}{\textbf{Electricity}} \\ \cmidrule(r){2-4}\cmidrule(lr){5-7}\cmidrule(l){8-10}
			~ & \multicolumn{1}{c}{MAE} & \multicolumn{1}{c}{RMSE} & \multicolumn{1}{c}{MAPE} & \multicolumn{1}{c}{MAE} & \multicolumn{1}{c}{RMSE} & \multicolumn{1}{c}{MAPE} & \multicolumn{1}{c}{MAE} & \multicolumn{1}{c}{RMSE} & \multicolumn{1}{c}{MAPE} \\ \midrule
			Mean & 104.63 & 169.79 & 164.75 & 45.24 & 61.20 & 37.56 & 2.98 & 4.64 & 86.63 \\
			TLE & 148.35 & 195.67 & 231.53 & 50.42 & 86.23 & 45.32 & 4.10 & 5.50 & 121.42 \\
			LO & 54.54 & 106.83 & 85.24 & 26.42 & 40.11 & 20.99 & 1.65 & 3.01 & 43.25 \\
			DCMF & 45.81 & 86.30 & 64.97 & 16.84 & 31.29 & 14.33 & 1.42 & 2.32 & 33.63 \\
			TRMF & 53.64 & 101.38 & 87.86 & 24.41 & 38.13 & 18.39 & 1.53 & 2.97 & 45.14 \\
			BRITS & 46.39 & 91.89 & 78.92 & 20.92 & 33.20 & 15.03 & 1.39 & 2.77 & 41.69 \\
			E\textsuperscript{2}GAN & 80.25 & 139.48 & 124.97 & 31.75 & 44.58 & 26.05 & 2.22 & 3.78 & 65.94 \\
			GRIN & $^*$38.53 & $^*$76.16 & $^*$49.94 & $^*$16.28 & $^*$29.60 & $^*$11.70 & $^*$1.26 & $^*$2.15 & $^*$25.93 \\
			CSDI & 44.78 & 93.48 & 59.29 & 18.39 & 33.94 & 14.98 & 1.36 & 2.42 & 38.94 \\
			mSSA & 50.39 & 96.93 & 84.29 & 22.39 & 35.59 & 17.10 & 1.49 & 2.99 & 44.88 \\
			\midrule
			NETS-ImpGAN & \underline{33.23} & \underline{66.43} & \underline{32.25} & \underline{15.78} & \underline{26.83} & \underline{10.04} & \underline{1.17} & \underline{1.98} & \underline{17.32} \\
			NETS-ImpGAN$^\dag$ & \textbf{30.38} & \textbf{64.18} & \textbf{31.14} & \textbf{13.27} & \textbf{24.45} & \textbf{9.29} & \textbf{1.06} & \textbf{1.65} & \textbf{15.23} \\
			\bottomrule
		\end{tabular}
	}
	\subtable[MV]{
		\begin{tabular}{@{}lrrrrrrrrr@{}}
			\toprule
			\multirow{2.5}*{\textbf{Method}} & \multicolumn{3}{c}{\textbf{Metro}} & \multicolumn{3}{c}{\textbf{Air}} & \multicolumn{3}{c}{\textbf{Electricity}} \\ \cmidrule(r){2-4}\cmidrule(lr){5-7}\cmidrule(l){8-10}
			~ & \multicolumn{1}{c}{MAE} & \multicolumn{1}{c}{RMSE} & \multicolumn{1}{c}{MAPE} & \multicolumn{1}{c}{MAE} & \multicolumn{1}{c}{RMSE} & \multicolumn{1}{c}{MAPE} & \multicolumn{1}{c}{MAE} & \multicolumn{1}{c}{RMSE} & \multicolumn{1}{c}{MAPE} \\ \midrule
			Mean & 101.36 & 156.39 & 158.49 & 44.23 & 56.23 & 37.24 & 2.81 & 4.25 & 81.58 \\
			TLE & 154.63 & 191.32 & 237.24 & 52.05 & 85.38 & 46.42 & 4.28 & 5.24 & 126.03 \\
			LO & 53.10 & 101.37 & 79.25 & 25.29 & 40.13 & 20.88 & 1.59 & 2.89 & 41.90 \\
			DCMF & 44.36 & 81.11 & 45.24 & 16.30 & 30.42 & 11.25 & 1.39 & 2.26 & 24.77 \\
			TRMF & 52.35 & 109.26 & 86.62 & 24.85 & 40.90 & 19.94 & 1.61 & 2.99 & 45.57 \\
			
			BRITS & 48.25 & 98.17 & 78.30 & 20.63 & 38.75 & 15.15 & 1.49 & 2.56 & 42.53 \\
			E\textsuperscript{2}GAN & 72.89 & 140.25 & 113.58 & 28.05 & 45.28 & 23.26 & 2.09 & 3.72 & 56.83 \\
			GRIN & $^*$44.35 & $^*$80.97 & $^*$44.78 & $^*$15.83 & $^*$30.17 & $^*$11.05 & $^*$1.33 & $^*$2.18 & $^*$23.90 \\
			CSDI & 46.93 & 92.36 & 57.11 & 18.29 & 36.39 & 14.39 & 1.50 & 2.67 & 43.92 \\
			mSSA & 51.03 & 103.59 & 80.93 & 23.49 & 39.61 & 18.76 & 1.59 & 2.97 & 44.24 \\
			\midrule
			NETS-ImpGAN & \underline{38.71} & \underline{62.46} & \underline{33.90} & \underline{14.46} & \underline{28.36} & \underline{10.55} & \underline{1.31} & \underline{2.05} & \underline{18.21} \\
			NETS-ImpGAN$^\dag$ & \textbf{35.36} & \textbf{59.09} & \textbf{32.74} & \textbf{13.04} & \textbf{25.99} & \textbf{9.69} & \textbf{1.20} & \textbf{1.77} & \textbf{15.74} \\
			\bottomrule
		\end{tabular}
	}
\end{table}

\begin{table}[h]
    \centering
    \caption{Prediction performance on Speed, whose missing rate is 30\%. The results of the baselines are the better of those refined by MIRACLE or not, and the results of {\model}$^\dag$ are refined by MIRACLE. \textbf{Bold}, \underline{underline} and superscript $^*$ indicate the best, second best and third best in each row.}
    \label{tbl:prediction_performance_speed}
    \renewcommand{\arraystretch}{0.95}
       \begin{tabular}{@{}l@{\hspace{1\tabcolsep}}r@{\hspace{1\tabcolsep}}c@{\hspace{1\tabcolsep}}c@{\hspace{1\tabcolsep}}c@{\hspace{1\tabcolsep}}c@{\hspace{1\tabcolsep}}c@{\hspace{1\tabcolsep}}c@{\hspace{1\tabcolsep}}c@{\hspace{1\tabcolsep}}c@{\hspace{1\tabcolsep}}c@{\hspace{1.4\tabcolsep}}c@{\hspace{1.5\tabcolsep}}c@{}}
            \toprule
            & {\small Mean} & {\small TLE} & {\small LO} & {\small DCMF} & {\small TRMF} & {\small BRITS} & {\small E\textsuperscript{2}GAN} & {\small GRIN} & {\small CSDI} & {\small mSSA} & {\small \specialcell{NETS-\\ImpGAN}} & {\small \specialcell{NETS-\\ImpGAN$^\dag$}} \\ \midrule
            
            MAE & 42.69 & 117.49 & 37.53 & 20.35 & 30.88 & 24.43 & 35.92 & $^*$16.69 & 21.70 & 27.49 & \underline{13.45} & \textbf{12.16} \\
            
            RMSE & 43.94 & 125.49 & 36.87 & 23.17 & 31.72 & 25.39 & 36.93 & $^*$18.75 & 22.48 & 28.70 & \underline{15.28} & \textbf{13.10} \\
            
            MAPE & 97.25 & 196.28 & 74.82 & 41.48 & 62.67 & 48.16 & 73.85 & $^*$37.74 & 43.36 & 53.84 & \underline{32.68} & \textbf{30.32} \\
            \bottomrule
        \end{tabular}
\end{table}

\subsection{Prediction Performance}\label{subsec:prediction_performance}

\subsubsection{Comparison with Baselines}\label{subsubsec:comparison_with_baselines}

We compare {\model} with the baselines under different missing patterns and missing rates. Table \ref{tbl:prediction_performance} shows the results on Metro, Air, Electricity under 25\% missing rate; those under the other settings are similar and omitted due to space limit. {\model} outperforms the baselines even when they are refined by MIRACLE, reducing the MAE of the second best GRIN by 2\%-16\%, with an average of 9\%, reducing the RMSE of GRIN by 6\%-22\%, with an average of 11\%, and reducing the MAPE of GRIN by 4\%-34\%, with an average of 20\%. Table \ref{tbl:prediction_performance_speed} shows the results on the naturally incomplete Speed dataset, whose missing rate is 30\%. The metrics are only computed on the observed future entries, since the ground truth of missing future is unavailable. {\model} outperforms the refined GRIN, reducing MAE by 21\%, RMSE by 17\% and MAPE by 15\%, which shows that {\model} can be applied to data with complex real-world missingness. When refined by MIRACLE, {\model} further reduces the MAE by 10\%, the RMSE by 8\% and MAPE by 6\% averaged over all four datasets. MIRACLE has similar level of improvement on the baselines and {\model}.

Note that Mean, TLE and LO have strong implicit assumptions on the data distribution. Mean assumes low temporal and inter-time series variation, TLE assumes linearity in the temporal dimension, and LO assumes low temporal variation. Since these assumptions hardly hold, these models have poor performance in general. In contrast, TRMF, BRITS, E\textsuperscript{2}GAN, CSDI and mSSA can capture temporal correlations without such restrictive assumptions and thus achieve better performance than Mean and TLE, but they do not exploit the underlying graph. DCMF and GRIN can further exploit the graph, leading to even better performance. However, DCMF relies on the assumed linear system model, GRIN has no direct control on the missing part of future, and its bidirectional architecture may not be a perfect match for prediction, which results in their worse performance than {\model}.

\subsubsection{Performance with Increasing Missing Rate}\label{subsubsec:performance_with_increasing_missing_rate}

\begin{figure}[th]
	\centering
	\subfigure[Random]{
% 		\begin{minipage}[t]{0.45\columnwidth}
			\centering
			\includegraphics[width=0.46\textwidth]{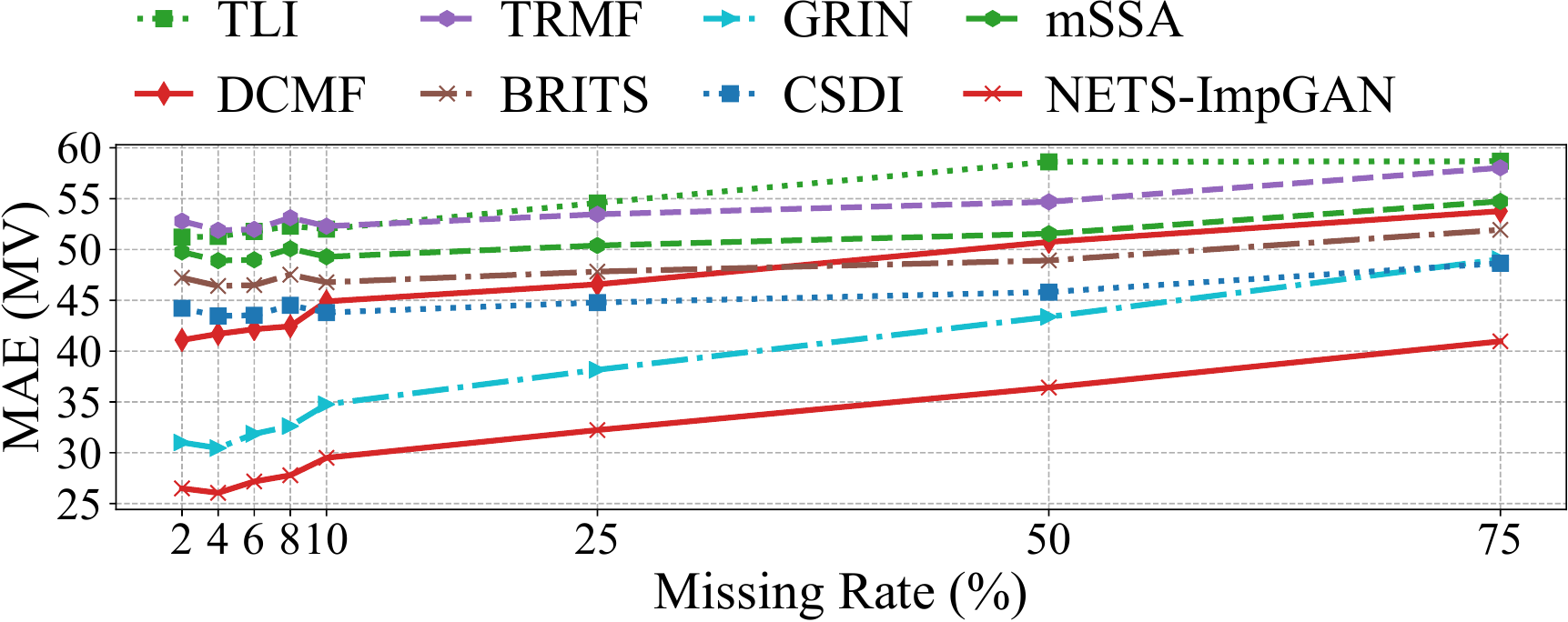}
% 		\end{minipage}
	}
	\hspace{1em}
	\subfigure[MV]{
% 		\begin{minipage}[t]{0.45\columnwidth}
			\centering
			\includegraphics[width=0.46\textwidth]{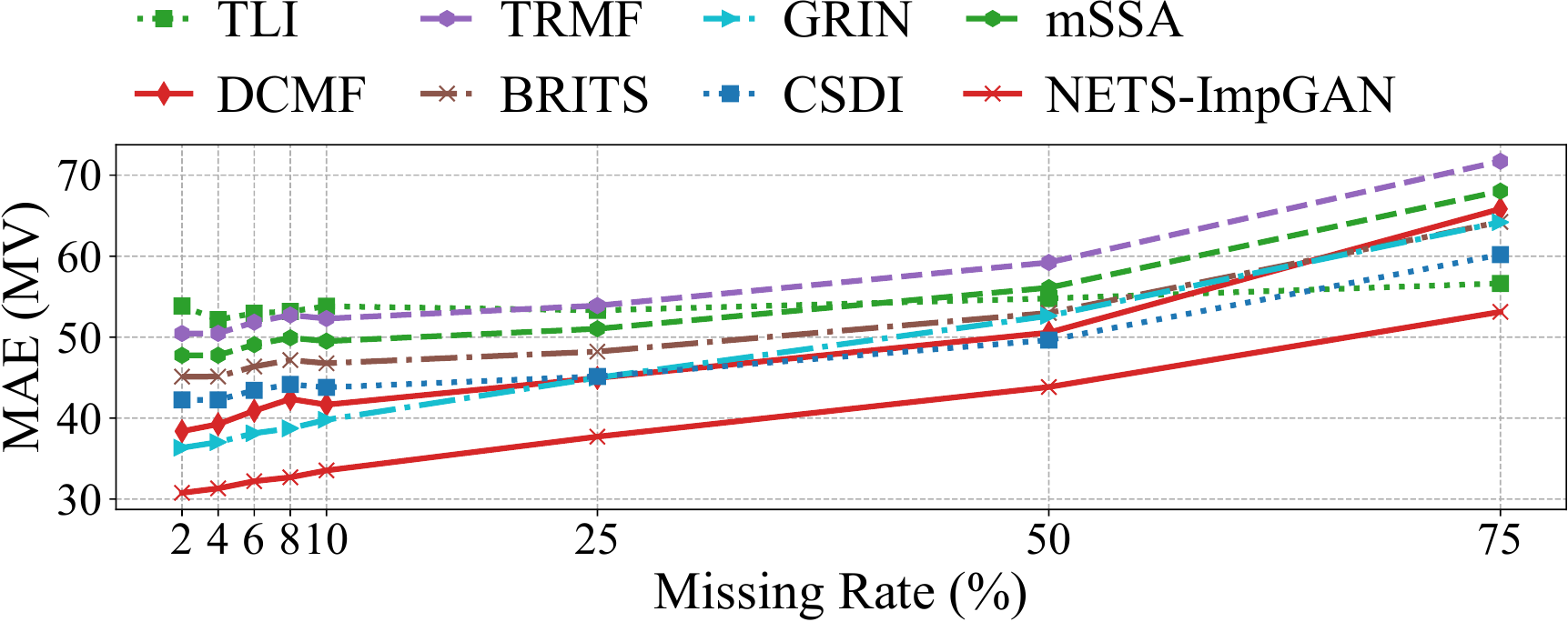}
% 		\end{minipage}
	}
	\caption{Prediction performance with increasing missing rate on Metro. The results of the baselines are the better of those refined by MIRACLE or not, and the results of {\model} is without refinement.}
	\Description{}
	\label{fig:prediction_with_increasing_missing_rate}
\end{figure}

Figure \ref{fig:prediction_with_increasing_missing_rate} shows the performance against missing rate on Metro. Since Mean, TLE and E\textsuperscript{2}GAN have much poorer performance than the others, we omit them in the figure for better visualization and easy comparison. Note that {\model} consistently outperforms the baselines across different missing rates and its performance degrades gracefully as missing rate increases.  The results on Air and Electricity are similar.

\subsubsection{Stepwise Performance}

\begin{figure}[thb]
	\centering
	\includegraphics[width=0.5\columnwidth]{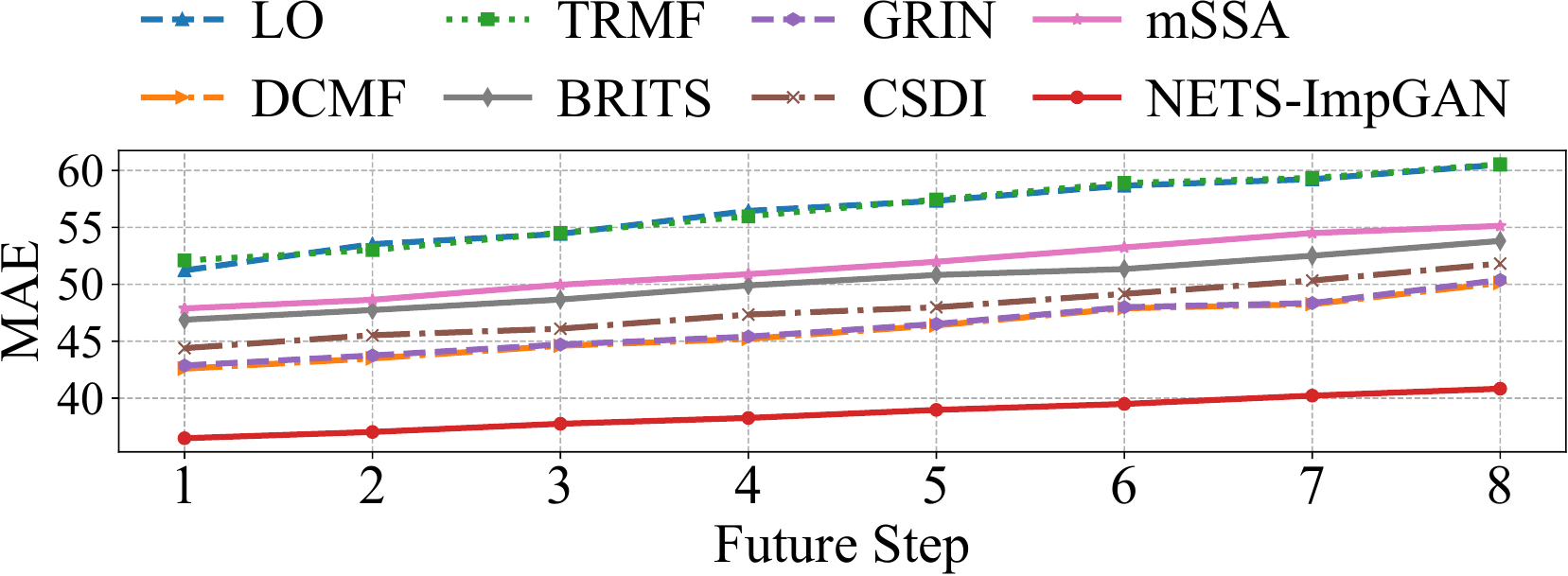}
	\caption{Stepwise prediction performance on Metro under MV and 25\% missing rate. The results of the baselines are the better of those refined by MIRACLE or not, and the results of {\model} is without refinement.}
	\Description{}
	\label{fig:multi-step_prediction}
\end{figure}

In this section, we study the prediction performance for individual future time steps rather than the overall performance. Figure \ref{fig:multi-step_prediction} shows the MAE on Metro under MV and 25\% missing rate against the number of steps into the future; those under the other settings are similar. We also omit Mean, TLE and E\textsuperscript{2}GAN. Note that {\model} has the best performance acoss all time steps. In particular, it performs much better than the baselines in long-term prediction.  Its accuracy for the eighth step is higher than those of the others even for the first step. Moreover, its performance degrades more slowly than the others as time step increases. This is partly because the convolution structure in GTAN helps avoid error accumulation typically suffered by recurrent structures in BRITS and DCMF.

\subsection{Comparison with Two-Phase Methods}\label{subsec:comparison_with_two-phase_prediction}

A common alternative way to deal with missing data is to use two-phase methods, which first impute incomplete data and then use the imputed data to train prediction models.
In this section, we demonstrate the advantage of {\model} over two-phase methods. To make things clear, in the rest of this section, we will distinguish the two usages of {\model}. It is referred to as {\model} when used for imputation, and {\modelp} for prediction, where the superscript star alludes to the mask $\bM^*$ used for prediction. We first show that {\model} achieves the state-of-the-art imputation performance. We then show that {\modelp} outperforms two-phase methods even when {\model} is used for imputation.

\subsubsection{Imputation Performance}\label{subsubsec:imputation_performance}

In this section, we evaluate the performance of {\model} for imputing the incomplete history. The baselines include Mean, DCMF, TRMF, BRITS, E\textsuperscript{2}GAN, GRIN, CSDI and mSSA, all of which are also used in Section \ref{subsec:prediction_performance}. We also add five more imputation baselines as follows: (1) simple methods including NA and TLI, and (2) state-of-the-art methods including WDGTC \cite{li2020tensor}, S-MKKM \cite{gong2020a} and SD-ADMM \cite{meyers2023signal}. NA and TLI are described below.
\begin{itemize}[leftmargin=*]
	\item \textbf{Neighborhood Average (NA)}. NA imputes a missing entry with the mean of observed values on its one-hop neighbors in the underlying graph at the same timestamp in the same sample. Mean is used if the required entries are missing.
	
	\item \textbf{Temporal Linear Interpolation (TLI)}. TLI imputes a missing entry with the mean of its last-observed historical and first-observed future entries on the same node in the same sample. Mean is used if the required entries are missing.
\end{itemize}
As we will use multiple imputation in two-phase prediction, in addition to MAE, we also measure the imputation quality using the Wasserstein Distance (WD) between the empirical distributions of real complete samples and imputed samples \cite{xu2018an}.

\begin{table}[t]
	\centering
	\caption{Differences between {\model} and imputation baselines.}
	\label{tbl:imputation_difference}
	\renewcommand{\arraystretch}{0.9}
	\begin{tabular}{@{}lccc@{}}
		\toprule
		& \specialcell{{Supervision with Incomplete Data}} & {Graph} & {Temporal} \\ \midrule
		{NA}		& $\times$	& $\surd$	& $\times$ \\
		{TLI}		& $\times$	& $\times$	& $\surd$ \\
		{WDGTC}		& $\times$	& $\surd$	& $\times$ \\
		{S-MKKM}    & $\times$	& $\surd$	& $\times$ \\
		{SD-ADMM}   & $\times$	& $\times$	& $\surd$ \\
		{{\model}}	& $\surd$	& $\surd$	& $\surd$ \\
		\bottomrule
	\end{tabular}
\end{table}

Table \ref{tbl:imputation_difference}  summarizes the differences between {\model} and the additional imputation baselines w.r.t. whether or not they can (1) supervise with incomplete data, (2) exploit the underlying graph, and (3) capture temporal correlations.

\begin{table}[t]
	\centering
	\caption{Imputation performance on Metro, Air and Electricity under 25\% missing rate. The results of the baselines are the better of those refined by MIRACLE or not, and the results of {\model}$^\dag$ are refined by MIRACLE.}
	\label{tbl:imputation_performance}
	\renewcommand{\arraystretch}{0.95}
	\subtable[Random]{
		\begin{tabular}{@{}lrrrrrrrr@{}}
			\toprule
			\multirow{2.5}*{\textbf{Method}} & \multicolumn{2}{c}{\textbf{Metro}} & \multicolumn{2}{c}{\textbf{Air}} & \multicolumn{2}{c}{\textbf{Electricity}} \\ \cmidrule(r){2-3}\cmidrule(lr){4-5}\cmidrule(l){6-7}
			~ & \multicolumn{1}{c}{MAE} & \multicolumn{1}{c}{WD} & \multicolumn{1}{c}{MAE} & \multicolumn{1}{c}{WD} & \multicolumn{1}{c}{MAE} & \multicolumn{1}{c}{WD} \\ \midrule
			Mean & 148.62 & 2870.13 & 65.38 & 1272.52 & 4.12 & 74.94 \\
			NA & 140.83 & 2251.52 & 47.17 & 769.99 & 3.95 & 57.26 \\
			TLI & 26.72 & 543.08 & 12.88 & 260.95 & 0.95 & 13.85 \\
			DCMF & 19.83 & 420.94 & 8.99 & 176.35 & 0.85 & 11.24 \\
			TRMF & 24.26 & 545.74 & 10.65 & 250.04 & 0.91 & 13.20 \\
			BRITS & 21.52 & 483.59 & 9.60 & 213.85 & 0.89 & 12.89 \\
			E\textsuperscript{2}GAN & 74.63 & 1598.37 & 29.24 & 613.40 & 2.08 & 40.25 \\
			WDGTC & 20.25 & 478.24 & 9.25 & 208.73 & 0.89 & 13.01 \\    
			S-MKKM & 21.50 & 484.23 & 9.63 & 219.20 & 0.95 & 14.96 \\
			GRIN & $^*$19.57 & $^*$413.60 & $^*$8.34 & $^*$153.57 & $^*$0.82 & $^*$10.17 \\
			CSDI & 20.93 & 479.72 & 9.18 & 201.13 & 0.87 & 12.65 \\
			SD-ADMM & 25.39 & 533.90 & 11.79 & 261.31 & 0.99 & 14.85 \\
			mSSA & 22.49 & 523.59 & 10.16 & 223.69 & 0.88 & 12.82 \\
			\midrule
			NETS-ImpGAN & \underline{17.03} & \underline{289.11} & \underline{8.22} & \underline{141.55} & \underline{0.76} & \underline{7.76} \\
			NETS-ImpGAN$^\dag$ & \textbf{16.36} & \textbf{257.33} & \textbf{8.17} & \textbf{133.70} & \textbf{0.70} & \textbf{6.58} \\
			\bottomrule
		\end{tabular}
	}
	\subtable[MV]{
		\begin{tabular}{@{}lrrrrrrrr@{}}
			\toprule
			\multirow{2.5}*{\textbf{Method}} & \multicolumn{2}{c}{\textbf{Metro}} & \multicolumn{2}{c}{\textbf{Air}} & \multicolumn{2}{c}{\textbf{Electricity}} \\ \cmidrule(r){2-3}\cmidrule(lr){4-5}\cmidrule(l){6-7}
			~ & \multicolumn{1}{c}{MAE} & \multicolumn{1}{c}{WD} & \multicolumn{1}{c}{MAE} & \multicolumn{1}{c}{WD} & \multicolumn{1}{c}{MAE} & \multicolumn{1}{c}{WD} \\ \midrule
			Mean & 172.97 & 2549.77 & 77.52 & 1125.75 & 4.74 & 62.74 \\
			NA & 118.42 & 2185.36 & 40.25 & 735.08 & 3.33 & 53.95 \\
			TLI & 27.99 & 569.15 & 13.28 & 246.17 & 1.19 & 14.59 \\
			DCMF & 21.43 & 399.41 & 9.06 & 189.23 & 0.91 & 13.57 \\
			TRMF & 25.74 & 537.86 & 11.72 & 280.69 & 1.12 & 15.30 \\
			BRITS & 23.47 & 460.58 & 10.15 & 222.98 & 0.96 & 13.92 \\
			E\textsuperscript{2}GAN & 75.32 & 1632.25 & 31.61 & 658.42 & 2.68 & 53.77 \\
			WDGTC & 21.88 & 503.71 & 10.40 & 238.56 & 1.05 & 14.70 \\
			S-MKKM & 22.47 & 491.44 & 9.46 & 213.57 & 1.09 & 15.34 \\
			GRIN & $^*$21.35 & $^*$407.16 & $^*$8.65 & $^*$163.08 & $^*$0.63 & $^*$6.93 \\
			CSDI & 21.94 & 466.69 & 9.39 & 245.86 & 0.70 & 7.49 \\
			SD-ADMM & 26.14 & 559.03 & 13.79 & 301.98 & 1.15 & 15.61 \\
			mSSA & 23.91 & 548.90 & 11.78 & 286.40 & 0.99 & 13.82 \\
			\midrule
			NETS-ImpGAN & \underline{18.17} & \underline{319.45} & \underline{8.07} & \underline{135.81} & \underline{0.51} & \underline{6.17} \\
			NETS-ImpGAN$^\dag$ & \textbf{17.79} & \textbf{307.50} & \textbf{7.78} & \textbf{121.39} & \textbf{0.47} & \textbf{5.88} \\
			\bottomrule
		\end{tabular}
	}
\end{table}

\begin{figure}[t]
	\centering
	\subfigure[Random]{
% 		\begin{minipage}[t]{0.7\columnwidth}
			\centering
			\includegraphics[width=0.46\textwidth]{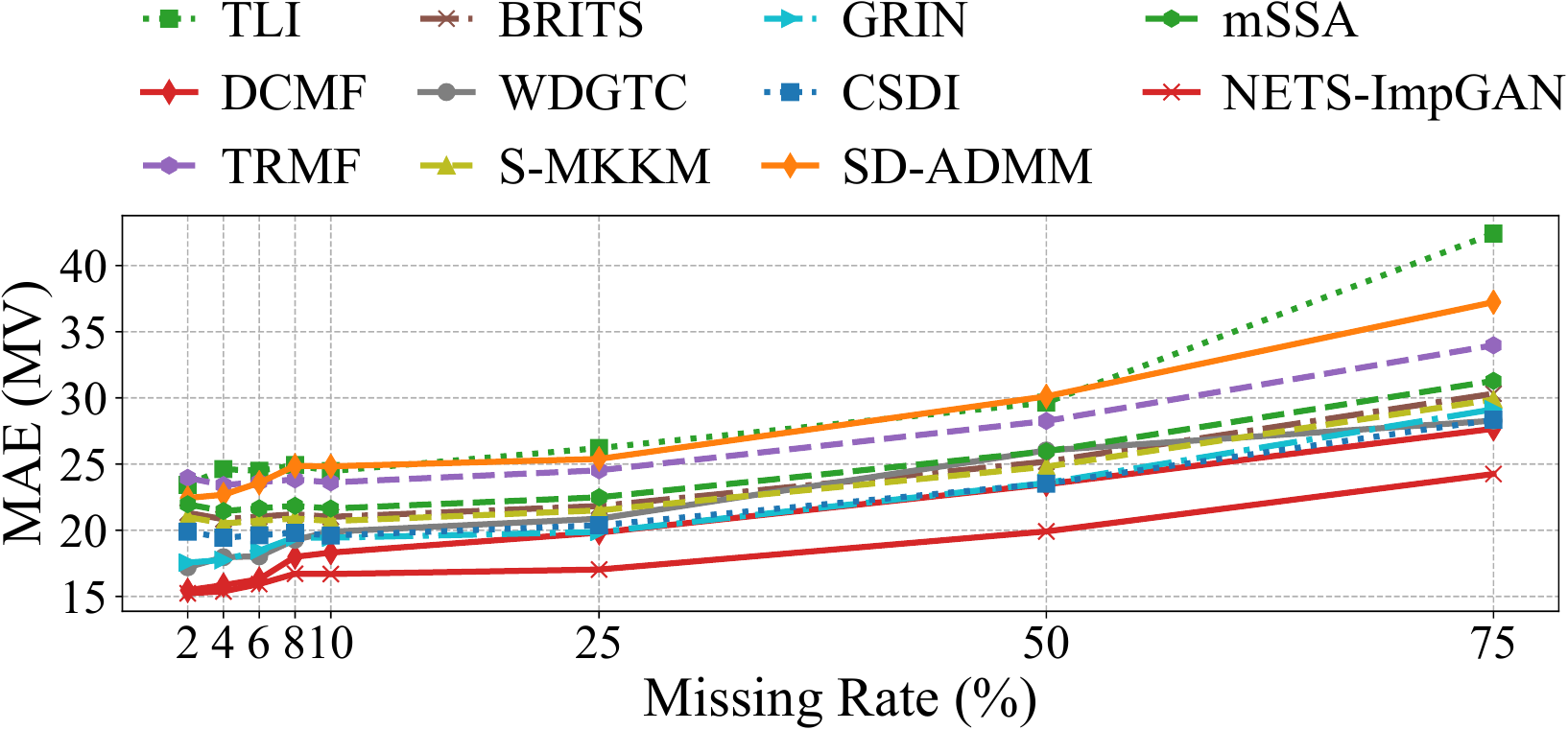}
% 		\end{minipage}
	}
	\hspace{1em}
	\subfigure[MV]{
% 		\begin{minipage}[t]{0.7\columnwidth}
			\centering
			\includegraphics[width=0.46\textwidth]{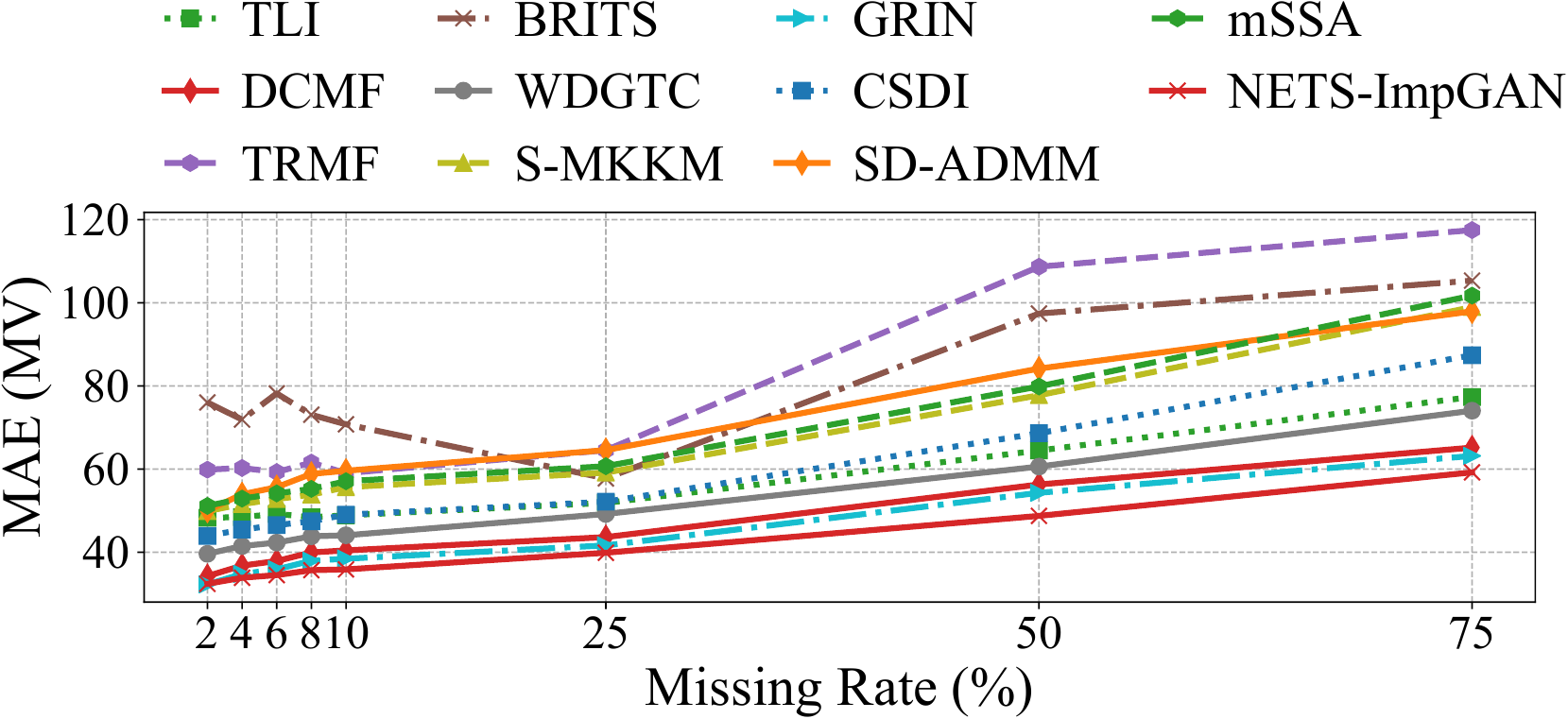}
% 		\end{minipage}
	}
	\caption{Imputation performance with increasing missing rate on Metro. The results of the baselines are the better of those refined by MIRACLE or not, and the results of {\model} is without refinement.}
	\Description{}
	\label{fig:imputation_with_increasing_missing_rate}
\end{figure}

Table \ref{tbl:imputation_performance} shows the results of single and multiple imputations on Metro, Air and Electricity under 25\% missing rate, where single imputation is measured by MAE and multiple imputation is measured by WD; those under the other missing rates are similar. Since the error on observed entries is always zero in the imputation task, we do not compare on Speed. {\model} achieves the best performance on all the datasets in both single and multiple imputations. Note that our reported results is different from those in \citet{cini2022filling} due to different experimental settings. Figure \ref{fig:imputation_with_increasing_missing_rate} shows the MAE of single imputation against missing rate on Metro; the other settings are similar. We omit Mean, NA and E\textsuperscript{2}GAN due to their poor performance. Similar to the prediction case, {\model} outperforms the baselines and the performance degrades gracefully with increasing missing rate. Note that the curves for BRITS bend up at very low missing rates due to overfitting. Comparison with baselines on downstream prediction task will be shown in Section \ref{subsubsec:two-phase_prediction}.

\subsubsection{Two-Phase Methods}\label{subsubsec:two-phase_prediction}

We compare {\modelp} with two-phase methods that use various combinations of imputation and prediction methods. For the prediction phase in two-phase methods, we use our {\imputationgenerator} and ten state-of-the-art methods that require complete data for training: those requiring complete input, including DCRNN \cite{li2018diffusion}, STGCN \cite{yu2018spatio-temporal}, Graph WaveNet \cite{wu2019graph} and LSGCN \cite{huang2020lsgcn}, and those do not, including STGNN-DAE \cite{kuppannagari2021spatio-temporal}, IGNNK \cite{wu2021inductive}, SSSDS4 \cite{alcaraz202diffusion-based}, RIHGCN \cite{zhong2021heterogeneous}, GSTAE \cite{wang2023traffic} and GENIE \cite{dockhorn2022genie}. RIHGCN can take incomplete history as input but requires complete future for supervision, thus we only impute the future part. Since RIHGCN is designed for NETS with multiple underlying graphs, we adapt its multi-graph module to single graph. Note that we do not select the methods that additionally require periodic or scenario-specific auxiliary information, since there is no such information in our setting even if the data is complete. For the imputation phase, we use all the methods in Section \ref{subsubsec:imputation_performance} and also include the commonly used zero imputation. Note that {\model} is used with both single and multiple imputation. The results are shown in Table \ref{tbl:two-phase_prediction}. Due to space limit, we omit those of STGCN, LSGCN, STGNN-DAE, SSSDS4 and RIHGCN that mostly have poorer performance.

\begin{table}[tp]
	\centering
	\caption{End-to-end vs. two-phase prediction on Metro under 25\% missing rate. Each row corresponds to an imputation method, and each column corresponds to a prediction method. The results of the baselines are the better of those refined by MIRACLE or not, and the results of {\model} is without refinement. \textbf{Bold} and \underline{underline} represent the best along each column and row, respectively. Our end-to-end prediction is shown at the bottom.}
	\label{tbl:two-phase_prediction}
	\renewcommand{\arraystretch}{0.95}
	\subtable[Random]{
		\begin{tabular}{@{}lcccccc@{}}
			\toprule
			 & {\small DCRNN} & {\small Graph WaveNet} & {\small IGNNK} & {\small GSTAE} & {\small GENIE} & {\small GTA U-Net} \\ \midrule
			{\small Zero} & 68.36 & 67.48 & 68.56 & 67.10 & 67.16 & \underline{65.32} \\ 
			{\small Mean} & 67.03 & 67.68 & 68.32 & 66.86 & 66.51 & \underline{64.53} \\ 
			{\small NA} & 66.52 & 65.02 & 65.67 & 63.99 & 64.25 & \underline{61.99} \\ 
			{\small TLI} & 62.45 & 61.63 & 62.10 & 61.20 & 62.35 & \underline{60.13} \\ 
			{\small DCMF} & 49.43 & 48.20 & 48.83 & 47.33 & 47.90 & \underline{46.24} \\
			{\small TRMF} & 58.43 & 59.13 & 59.35 & 59.15 & 59.23 & \underline{57.25} \\ 
			{\small BRITS} & 53.67 & 53.01 & 53.88 & 53.17 & 53.28 & \underline{52.16} \\
			{\small E\textsuperscript{2}GAN} & 64.89 & 64.12 & 64.57 & 63.50 & 63.28 & \underline{61.89} \\
			{\small WDGTC} & 46.57 & 45.34 & 46.32 & 45.26 & 45.11 & \underline{43.39} \\
			{\small S-MKKM} & 54.39 & 54.92 & 55.90 & 54.86 & 56.23 & \underline{53.91} \\
			{\small GRIN} & 41.08 & 41.09 & 41.53 & 41.72 & 41.56 & \underline{39.16} \\
			{\small CSDI} & 43.48 & 42.29 & 44.56 & 42.55 & 43.78 & \underline{41.75} \\
			{\small SD-ADMM} & 58.35 & 57.60 & 58.38 & 57.77 & 58.95 & \underline{56.25} \\
			{\small mSSA} & 55.19 & 54.93 & 55.28 & 54.66 & 56.17 & \underline{54.02} \\
			{\small NETS-ImpGAN (Single)} & 39.42 & 38.97 & 39.08 & 38.77 & 39.01 & \underline{38.32} \\
			{\small NETS-ImpGAN (Multiple)} & \textbf{38.02} & \textbf{37.62} & \textbf{39.03} & \textbf{38.72} & \textbf{38.80} & \underline{\textbf{37.17}} \\ \midrule
			{\small \modelp } & \multicolumn{6}{c}{\textbf{33.23}} \\
			\bottomrule
		\end{tabular}
	}
	\subtable[MV]{
		\begin{tabular}{@{}lcccccc@{}}
			\toprule
			& {\small DCRNN} & {\small Graph WaveNet} & {\small IGNNK} & {\small GSTAE} & {\small GENIE} & {\small GTA U-Net} \\ \midrule
			{\small Zero} & 69.75 & 67.76 & 68.13 & 67.79 & 67.65 & \underline{66.45} \\
			{\small Mean} & 69.16 & 66.67 & 66.53 & 66.89 & 66.77 & \underline{65.42} \\
			{\small NA} & 66.79 & 65.23 & 65.17 & 64.29 & 64.88 & \underline{62.01} \\
			{\small TLI} & 63.35 & 60.97 & 61.73 & 60.98 & 61.13 & \underline{60.02} \\
			{\small DCMF} & 48.43 & 47.03 & 47.14 & 47.20 & 47.29 & \underline{46.24} \\
			{\small TRMF} & 60.75 & 59.25 & 60.69 & 59.16 & 59.33 & \underline{57.20} \\
			{\small BRITS} & 56.67 & 55.57 & 56.75 & 55.23 & 55.39 & \underline{54.45} \\
			{\small E\textsuperscript{2}GAN} & 64.03 & 64.46 & 65.23 & 63.59 & 63.29 & \underline{60.88} \\
			{\small WDGTC} & 47.10 & 44.95 & 46.11 & 44.47 & 44.91 & \underline{44.26} \\
			{\small S-MKKM} & 56.93 & 55.38 & 56.48 & 55.32 & 56.25 & \underline{55.70} \\
			{\small GRIN} & 46.58 & 44.56 & 45.73 & 44.97 & 44.95 & \underline{44.36} \\
			{\small CSDI} & 48.35 & 47.62 & 49.03 & 48.01 & 48.95 & \underline{46.80} \\
			{\small SD-ADMM} & 59.71 & 59.24 & 60.44 & 59.37 & 59.88 & \underline{58.69} \\
			{\small mSSA} & 57.61 & 57.30 & 59.42 & 57.98 & 58.78 & \underline{56.25} \\
			{\small NETS-ImpGAN (Single)} & 45.26 & 44.27 & 45.61 & 44.19 & 44.89 & \underline{43.95} \\
			{\small NETS-ImpGAN (Multiple)} & \textbf{43.90} & \textbf{43.72} & \textbf{44.50} & \textbf{43.27} & \textbf{44.10} & \underline{\textbf{42.81}} \\ \midrule
			{\small \modelp} & \multicolumn{6}{c}{\textbf{38.71}} \\
			\bottomrule
		\end{tabular}
	}
\end{table}

Among the prediction methods, the variants using {\imputationgenerator}, Graph WaveNet, GSTAE and GENIE have similar performance, slightly better than those of DCRNN and IGNNK; among the imputation methods, the variants using {\model} outperforms the others. Multiple imputation performs slightly better than single imputation, potentially because it better reflects the uncertainty in imputation. However, all the two-phase methods are outperformed by {\modelp}. Note that the combination {\model} plus {\imputationgenerator} have the same capability of capturing inter-time series and temporal correlations as {\modelp}, but the latter allows training with incomplete data in an end-to-end manner and hence avoids error accumulation in two-phase methods.

\subsection{Efficacy Study}\label{subsec:efficacy_study}

\subsubsection{Ablation Study}\label{subsubsec:ablation_study}

We consider the following ablated variants.
\begin{itemize}[leftmargin=*]
	\item {\model}-w/o-GAT. We remove all the GAT modules in {\neuralnetwork}s.
	\item {\model}-w/o-SA. We remove all the Multi-Head Self-Attention modules in {\neuralnetwork}s.
	\item {\model}-w/o-T-Conv. We remove all the T-Conv modules in {\neuralnetwork}s.
	\item {\model}-w/o-Atten. We remove all the attention mechanisms. We replace GAT by GraphConv \cite{morris2019weisfeiler}, which is designed for graphs with static weights. We set the edge weights for GraphConv \cite{morris2019weisfeiler} as follows. We retain the transition probability and processed distance as edge weight for Metro and Air, respectively, and set all edge weights to be 1 for Electricity and Speed, which is similar to the graph construction in previous works \cite{ou2020spt-trellisnets,li2018diffusion,cini2022filling,zhang2017deep}. We also remove all the Multi-Head Self-Attention modules.
\end{itemize}

Table \ref{tbl:ablation_study} shows the MAE for prediction on Metro under 25\% missing rate. {\model} achieves the best performance. Compared to the variants without GAT, Multi-Head Self-Attention, T-Conv and attention mechanism, the full model {\model} on average reaches 23\%, 8\%, 65\% and 15\% improvement respectively,
which affirms the efficacy of these components. Note the particular importance of the T-Conv, without which the performance deteriorates significantly.

\begin{table}[t]
	\centering
	\caption{{\model} vs. ablated variants on Metro under 25\% missing rate. Results are without refinement.}
	\label{tbl:ablation_study}
	\renewcommand{\arraystretch}{0.8}
	\begin{tabular}{@{}lccccc@{}}
		\toprule
		& \specialcell{-w/o-T-Conv} & \specialcell{-w/o-GAT} & \specialcell{-w/o-SA} & \specialcell{-w/o-Atten} & full NETS-ImpGAN \\ \midrule
		Random & 140.97 & 43.37 & 38.86 & 41.44 & \textbf{33.23} \\ \midrule
		MV & 84.12 & 48.59 & 40.98 & 43.22 & \textbf{38.71} \\
		\bottomrule
	\end{tabular}
\end{table}

\subsubsection{Comparison with Alternative Frameworks}\label{subsubsec:comparison_with_misgan}

We study the efficacy the {\framework} framework by comparing {\framework} with alternative frameworks, including GAIN \cite{yoon2018gain}, MisGAN \cite{li2019learning}, Partial VAE \cite{ma2019eddi}, MIWAE \cite{mattei2019miwae} and P-BiGAN \cite{li2020learning}. For a fair comparison, we use the same {\neuralnetwork}s inside these frameworks, so that they have the same capability of capturing inter-time series and temporal correlations as {\model}. The resulted models are denoted by prefix ``NETS-''.

Table \ref{tbl:framework_comparison} shows the MAE for prediction on Metro under 25\% missing rate. The results show that {\framework} outperforms the other frameworks, which demonstrates the efficacy of {\framework}.

\begin{table}[t]
	\centering
	\caption{{\framework} vs. alternative frameworks on Metro under 25\% missing rate. Results are without refinement.}
	\label{tbl:framework_comparison}
	\renewcommand{\arraystretch}{0.8}
	\begin{tabular}{@{}lcccccc@{}}
		\toprule
		& \specialcell{NETS-\\GAIN} & \specialcell{NETS-\\MisGAN} & \specialcell{NETS-\\Partial VAE} & \specialcell{NETS-\\MIWAE} & \specialcell{NETS-\\P-BiGAN} & \specialcell{{\model}} \\ \midrule
		Random & 44.56 & 38.95 & 40.60 & 42.71 & 38.85 & \textbf{33.23} \\ \midrule
		MV &  52.18 & 46.77 & 49.24 & 45.98 & 43.07 & \textbf{38.71} \\
		\bottomrule
	\end{tabular}
\end{table}

We further show that {\framework} avoids the issue of error accumulation in MisGAN and is better suited for imputation/prediction. We compare {\model} with NETS-MisGAN and two additional variants as follows.
\begin{itemize}[leftmargin=*]
	\item NETS-MisGAN-RealData. It  trains the imputer of MisGAN using real complete samples instead of generated ones, which is also equivalent to training {\model} with zero missing rate.
	\item {\model}-RealMask. It uses real mask samples in place of those generated by $G_m$.
\end{itemize}

Table \ref{tbl:comparison_with_misgan} shows the MAE for prediction on Metro under 25\% missing rate. Note that {\model} outperforms NETS-MisGAN by a large margin. The large performance gap between NETS-MisGAN-RealData and NETS-MisGAN indicates significant error accumulation in MisGAN due to the imperfectly learned joint data distribution. This issue is avoided by {\model}.
The imperfectly learned mask distribution can also cause error accumulation, but a comparison between {\model} and {\model}-RealMask shows this effect is negligible. This is because it is easy to learn the binary mask distribution, whose samples are fully observed.

\begin{table}[t]
	\centering
	\caption{{\model} vs. NETS-MisGAN on Metro under 25\% missing rate. Results are without refinement.}
	\label{tbl:comparison_with_misgan}
	\renewcommand{\arraystretch}{0.8}
	\begin{tabular}{@{}lcccc@{}}
		\toprule
		& \specialcell{\small NETS-MisGAN} & \specialcell{\small NETS-MisGAN-RealData} & \specialcell{\small NETS-ImpGAN} & \specialcell{\small NETS-ImpGAN-RealMask} \\ \midrule
		Random & 38.95 & 29.44 & 33.23 & 33.32 \\ \midrule
		MV & 46.77 & 31.31 & 38.71 & 38.52 \\
		\bottomrule
	\end{tabular}
\end{table}

\subsubsection{Visualization of GTAN Components}\label{subsubsec:attention_visualization}

In this section, we give illustrative examples of GAT, Multi-Head Self-Attention and T-Conv respectively.

\begin{figure}[t]
	\centering
	\subfigure[GAT]{
		\begin{minipage}[t]{0.8\columnwidth}
			\centering
			\includegraphics[width=0.8\textwidth]{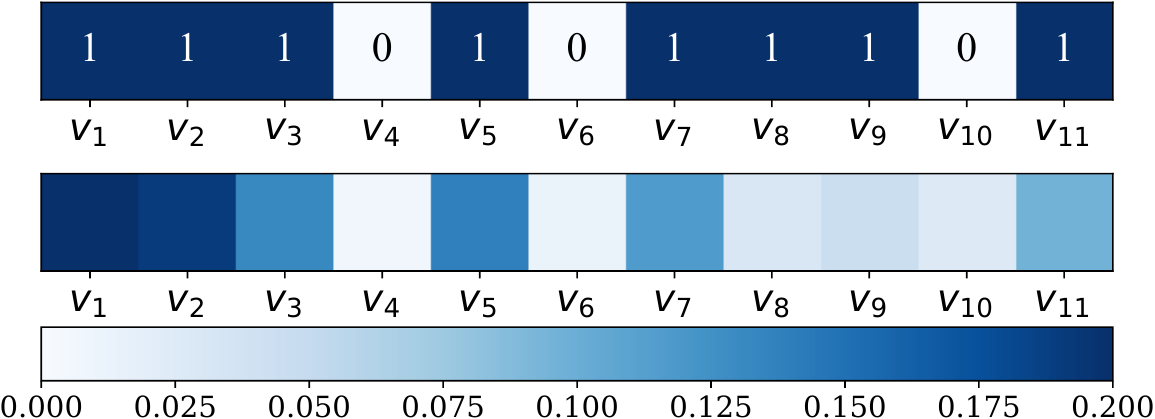}
			\label{fig:attention_visualization_gat}
		\end{minipage}
	}
	
	\subfigure[Multi-Head Self-Attention]{
		\begin{minipage}[t]{0.8\columnwidth}
			\centering
			\includegraphics[width=.9\textwidth]{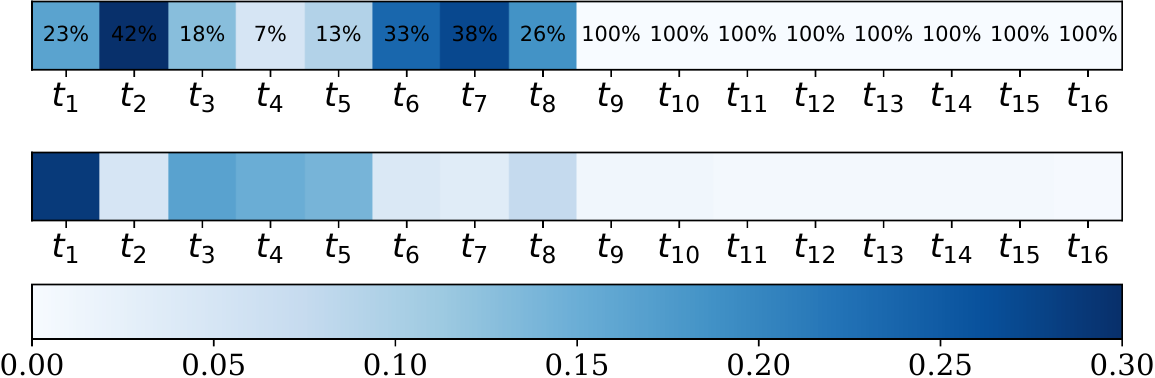}
			\label{fig:attention_visualization_sa}
		\end{minipage}
	}
	
	\subfigure[T-Conv]{
		\begin{minipage}[t]{0.8\columnwidth}
			\centering
			\includegraphics[width=.9\textwidth]{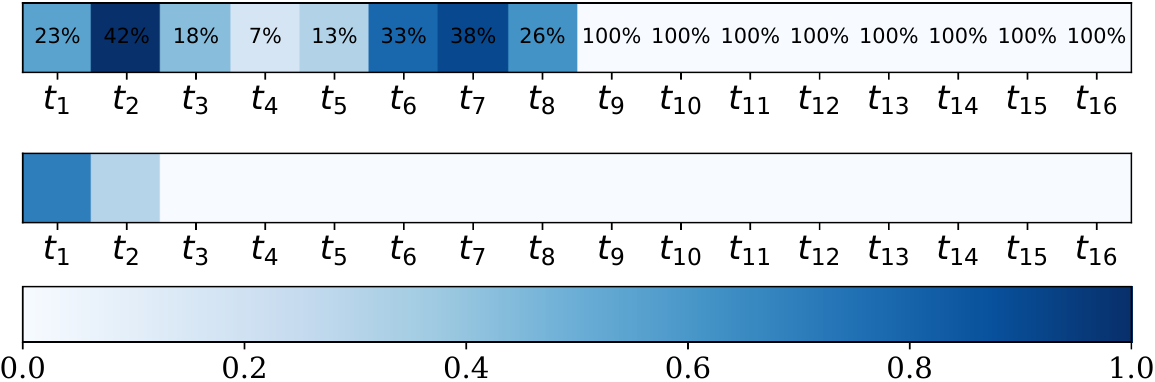}
			\label{fig:visualization_tconv}
		\end{minipage}
	}
	\caption{Visualization of GTAN components.}
    \label{fig:attention_visualization}
\end{figure}

Figure \ref{fig:attention_visualization_gat} visualizes the learned attention coefficients of GAT for one node in a sample. The node has 11 neighboring nodes, denoted by $v_1, \dots, v_{11}$, the mask of which is indicated by the first line. The corresponding attention coefficients of these nodes are shown in the second line, with deeper color representing larger coefficient. We can observe that more attention are assigned to nodes with observed values, which suggests that the attention mechanism helps differentiating the observed values and noise.

Figure \ref{fig:attention_visualization_sa} visualizes an example of learned attention coefficients of self attention. Note that a sample consist of 16 timestamps; for each of the first 8 historical timestamps, some nodes are observed and the others are missing, and the last 8 future timestamps are all treated as missing. In this example, we show the attention paid by the first timestamp in the output to the 16 timestamps in the input. The missing rate of each timestamp is indicated by the first line, and the corresponding attention coefficients are shown in the second line. We can observe that more attention are assigned to timestamps with lower missing rate, except that the highest attention is assigned to the first timestamp in the input, since a timestamp usually has very strong correlation with itself. The visualization also demonstrates that the attention mechanism can help pay more attention to observed values.

Figure \ref{fig:visualization_tconv} visualizes the normalized learned weights of the kernel of size $3$ in the first T-Conv layer. In this example, we show the kernel corresponding to the first timestamp in the output. Since T-Conv captures temporal correlations from a local view, this first output timestamp only depends on the first and second input timestamp. The weights at $t_1$ and $t_2$ are greater than zero, while the others are zero. In contrast, all the attention coefficients in Figure \ref{fig:attention_visualization_sa} are greater than zero.

\subsection{Efficiency Study}\label{subsec:efficiency_study}

\subsubsection{Data Efficiency}\label{subsubsec:data_efficiency}

Given a dataset consisting of both complete and incomplete samples, one way to apply methods that require complete data for training is to simply remove all the incomplete samples, which could lead to low data efficiency as mentioned in Section \ref{sec:introduction}. In this section, we take a closer look at this issue. We will refer to  methods that require complete data for training simply as complete-data methods.  We 
mainly focus on the following complete-data methods that do not require complete input: STGNN-DAE, IGNNK, SSSDS4, RIHGCN, GSTAE and GENIE. We do not consider those requiring complete input to exclude the influence of zero-imputed history on the performance when they are fed with incomplete history. For the experiment, we manually introduce missing values in a training set consisting of only complete samples initially, and vary the proportion of incomplete samples. We train {\model} with the entire training set, consisting of both complete and incomplete samples, and train the complete-data methods with only the complete samples. Note that for complete-data methods, the complete samples are used as ground truth, and the actual training inputs are generated by randomly masking the ground truth using the target missing pattern and missing rate, both assumed to be known. All the methods are tested under the same missing pattern and missing rate as the incomplete samples.

\begin{figure}[t]
	\centering
	\subfigure[Removing incomplete samples (25\% missing rate).]{
% 		\begin{minipage}[t]{0.5\columnwidth}
			\centering
			\includegraphics[width=.46\textwidth]{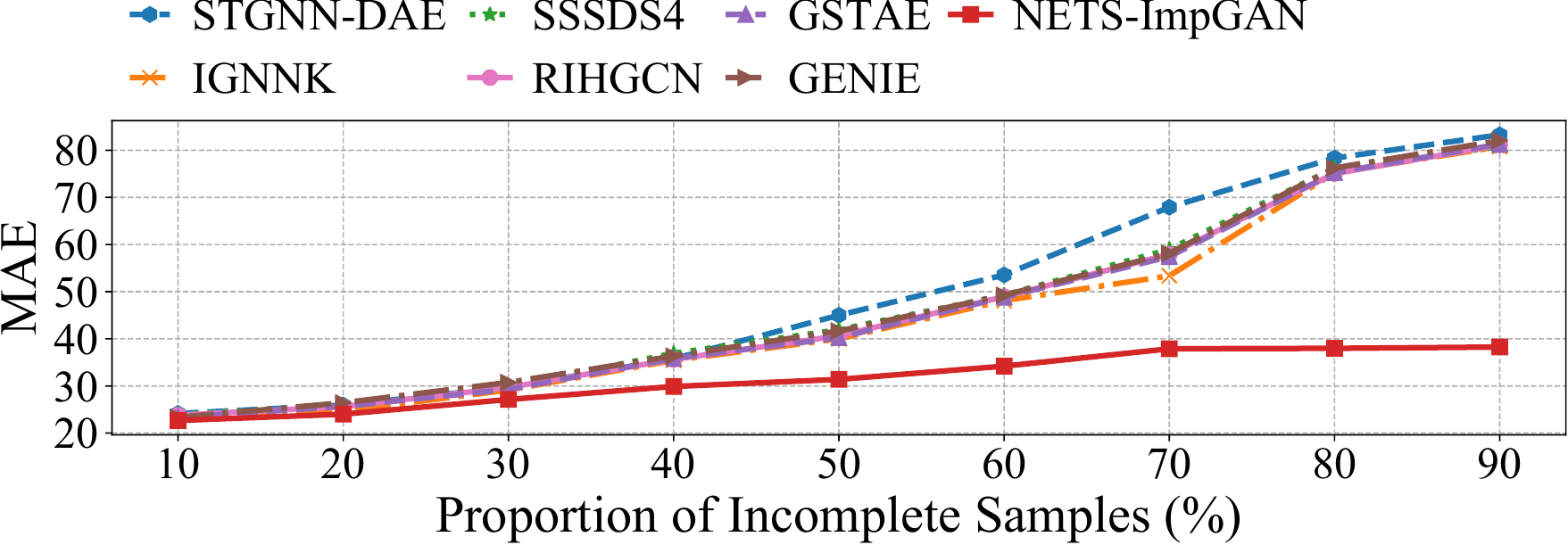}
			\label{fig:partial_complete_remove}
% 		\end{minipage}
	}
	\hspace{1em}
	\subfigure[Zero imputation (2\% missing rate).]{
% 		\begin{minipage}[t]{0.5\columnwidth}
			\centering
			\includegraphics[width=.46\textwidth]{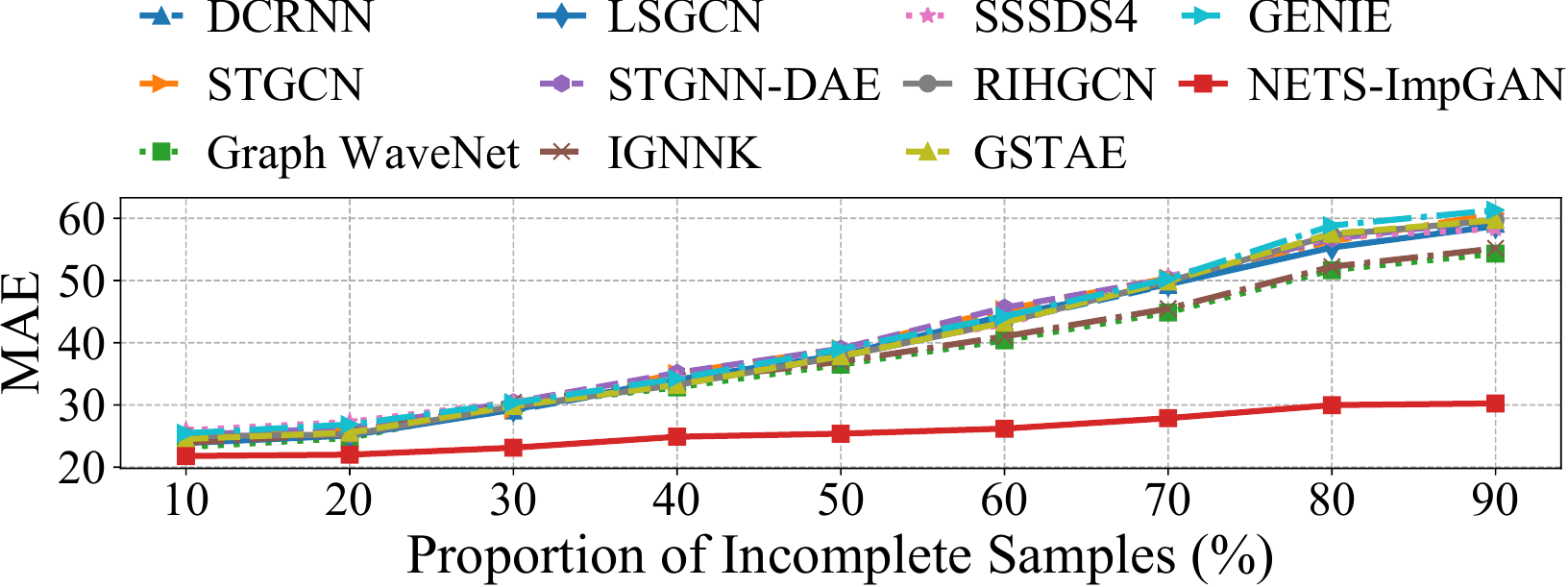}
			\label{fig:partial_complete_zero}
% 		\end{minipage}
	}
	\caption{Performance on partially complete Metro dataset under MV.}
	\Description{}
    \label{fig:partial_complete}
\end{figure}

Figure \ref{fig:partial_complete_remove} shows the results on Metro, where the incomplete part follows MV and 25\% missing rate and the proportion of incomplete samples varies from 10\% and 90\% at a step of 10\%. 
When the proportion of incomplete samples is small, the complete-data methods and {\model} have similar performance, as there are enough complete samples. As the proportion of incomplete samples increases, the performance gap between the complete-data methods and {\model} becomes larger, as only {\model} can exploit the incomplete samples. This shows that removing incomplete samples can lead to low data efficiency and poor performance.

Another common practice to exploit incomplete samples in complete-data methods is to first impute the missing entries with zeros, and then use both imputed and originally complete samples for training \cite{zhong2021heterogeneous}. We also explore the performance of such adaptation on partially complete dataset. Here we additionally consider the following methods that require complete input: DCRNN, STGCN, Graph WaveNet and LSGCN. Figure \ref{fig:partial_complete_zero} shows the results under the same setting as in Figure \ref{fig:partial_complete_remove}, except that we set the missing rate of the incomplete part to the very low level of 2\% in favor of zero imputation. The training procedure is also the same, except that the ground truth set is now obtained by zero imputing the original incomplete samples. 
All the methods are tested under MV and 2\% missing rate, and we also use zero imputation for those requiring complete input.
While zero imputation improves on removing incomplete samples, the gap between the complete-data methods and {\model} still becomes large with the increase of the incomplete proportion.

\subsubsection{Model Efficiency}

In this section, we compare the efficiency of {\model} with that of the baselines in terms of time cost and model complexity. Table \ref{tbl:time_cost} shows the training time and testing time per sample on Metro under MV and 25\% missing rate. Mean, TLE and LO have very short testing time but in general poor performance, thus we omit them in the table.

\begin{table}[t]
	\centering
	\caption{Time cost on Metro under MV and 25\% missing rate.}
	\label{tbl:time_cost}
	\renewcommand{\arraystretch}{0.8}
	\begin{tabular}{@{}l@{\hspace{1\tabcolsep}}c@{\hspace{1\tabcolsep}}c@{\hspace{1\tabcolsep}}c@{\hspace{1\tabcolsep}}c@{\hspace{1\tabcolsep}}c@{\hspace{1\tabcolsep}}c@{\hspace{1\tabcolsep}}c@{\hspace{1\tabcolsep}}c@{}}
		\toprule
		& DCMF & TRMF & BRITS & E\textsuperscript{2}GAN & GRIN & CSDI & mSSA & \specialcell{NETS-\\ImpGAN} \\ \midrule
		\specialcell{Training Time (hours)} & 0 & 0 & 1.12 & 1.25 & 8.16 & 2.33 & 0 & 9.85 \\ \midrule
		\specialcell{Testing Time (seconds/sample)} & 0.49 & 0.37 & 0.32 & 1.08 & 0.26 & 0.24 & 0.52 & 0.19 \\
		\bottomrule
	\end{tabular}
\end{table}

Among these methods, DCMF, TRMF and mSSA do not need training but requires more time than most of the others for each testing sample respectively.
In contrast, deep learning methods, including BRITS, E\textsuperscript{2}GAN, GRIN, CSDI and {\model}, require a long time for training, but typically less time for testing, except for E\textsuperscript{2}GAN, which takes even more time than DCMF, TRMF and mSSA for testing.
Compared to BRITS and GRIN that involve recurrent computation, our {\model} uses convolution modules that enjoy concurrent computation, leading to even faster testing process. From the time cost perspective, {\model} is better suited for scenarios where the model does not need to be updated frequently and the prediction accuracy is more important, while DCMF, TRMF and mSSA are better suited for scenarios where the model needs to be updated frequently but is used to predict only a small number of times after each update.

In terms of model complexity, DCMF, TRMF and mSSA have a small number of parameters, as the former assumes a somewhat restrictive linear system model, and the latter is based on matrix factorization. Among the deep learning methods, BRITS, E\textsuperscript{2}GAN and CSDI have fewer parameters than {\model}, as the former three only have temporal modules. GRIN and {\model} have a similar complexity. Note that none of the smaller models can increase their number of parameters in a straightforward way to improve their performance.
%%!TEX root = ../main.tex

\section{Conclusion}\label{sec:conclusion}

In this paper, we study the \textit{prediction of NETS with incomplete data}.
We propose \textit{\model}, a novel deep learning framework for both prediction and imputation that can be trained on incomplete data with missing values in both history and future. We design \textit{Graph Temporal Attention Networks} that can adapt to different samples when capturing the inter-time series and temporal correlations. We conduct extensive experiments on four real-world datasets under different missing patterns and missing rates.
The results show that {\model} outperforms existing methods,
reducing the MAE of the second best up to 25\%.

\newpage
\bibliographystyle{ACM-Reference-Format}
\bibliography{contents/bib}

\end{document}